\documentclass[letterpaper, 10 pt, journal, twoside]{IEEEtran}
\usepackage[utf8]{inputenc} 
\usepackage[T1]{fontenc}    
\usepackage{url}            
\usepackage{booktabs}       
\usepackage{amsfonts}       
\usepackage{nicefrac}       
\usepackage{microtype}      
\usepackage{xcolor}         
\usepackage{xfrac}
\usepackage{makecell}
\usepackage[hidelinks]{hyperref}

\usepackage[style=ieee,backend=biber,url=false,doi=false,natbib=true,mincitenames=1,dashed=false,isbn=false]{biblatex}
\AtEveryBibitem{\clearfield{issn}}
\AtEveryCitekey{\clearfield{issn}}

\usepackage{amsmath}
\usepackage{amssymb}
\usepackage{mathtools}

\usepackage{enumitem}
\usepackage[detect-all=true]{siunitx}
\usepackage{tikz}
\usepackage{float}

\usepackage{xcolor}         
\usepackage{caption}
\usepackage{subcaption}

\usepackage{ragged2e}
\newcolumntype{L}[1]{>{\RaggedLeft\arraybackslash}p{#1}}
\newcolumntype{P}[1]{>{\centering\arraybackslash}p{#1}}

\renewrobustcmd{\bfseries}{\fontseries{b}\selectfont}
\renewrobustcmd{\boldmath}{}

\newrobustcmd{\B}{\bfseries} 
\usepackage{numprint}

\usepackage[capitalize,noabbrev]{cleveref}

\newcommand{\R}{\mathbb{R}}

\newcommand{\E}{\mathop{\mathbb{E}}}

\newcommand{\ours}{dG-VAE}

\usepackage{color,soul}

\usepackage{pgfplots}
\pgfplotsset{compat=newest}
\definecolor{pastelMagenta}{HTML}{FF48CF} 
\definecolor{pastelPurple}{HTML}{8770FE} 
\definecolor{pastelBlue}{RGB}{0,114,178} 
\definecolor{pastelSkyBlue}{RGB}{86,180,233} 
\definecolor{pastelSeaGreen}{RGB}{86,180,233} 
\definecolor{pastelGreen}{RGB}{0,158,115} 
\definecolor{pastelOrange}{RGB}{230,159,0} 
\definecolor{pastelRed}{HTML}{F5615C} 
\definecolor{darkColor}{HTML}{300A24} 
\pgfplotscreateplotcyclelist{pastelcolors}{%
	solid, pastelPurple, mark=none\\%
	solid, pastelBlue, mark=none\\%
	solid, pastelGreen, mark=none\\%
	solid, pastelRed, mark=none\\%
	solid, pastelMagenta, mark=none\\%
	solid, pastelOrange, mark=none\\%
	solid, pastelSkyBlue, mark=none\\%
}

\begin{document}

\title{Disentangled Neural Relational Inference for \\ Interpretable Motion Prediction}

\author{Victoria M. Dax$^{1}$, Jiachen Li$^{1}$, Enna Sachdeva$^{2}$, Nakul Agarwal$^{2}$, and Mykel J. Kochenderfer$^{1}$
    \thanks{Manuscript received: July, 23, 2023; Revised October, 17, 2023; Accepted November, 13, 2023.}
    \thanks{This paper was recommended for publication by Aleksandra Faust upon evaluation of the Associate Editor and Reviewers' comments. This work was supported by Honda Research Institute, USA}
    \thanks{$^{1}$V. M. Dax, J. Li, and M. J. Kochenderfer are with the Stanford Intelligent Systems Laboratory (SISL), Stanford University, USA. {\tt\small \{vmdax, jiachen\_li,mykel\}@stanford.edu}.}
    \thanks{$^{2}$E. Sachdeva and N. Agarwal are with the Honda Research Institute, USA. {\tt\small \{enna\_sachdeva, nakul\_agarwal\}@honda-ri.com}}
    \thanks{Digital Object Identifier (DOI): see top of this page.}
}

\markboth{IEEE Robotics and Automation Letters. Preprint Version. Accepted November, 2023}
{Dax \MakeLowercase{\textit{et al.}}: Disentangled Neural Relational Inference for Interpretable Motion Prediction}

\maketitle

\begin{abstract}
Effective interaction modeling and behavior prediction of dynamic agents play a significant role in interactive motion planning for autonomous robots. Although existing methods have improved prediction accuracy, few research efforts have been devoted to enhancing prediction model interpretability and out-of-distribution (OOD) generalizability. This work addresses these two challenging aspects by designing a variational auto-encoder framework that integrates graph-based representations and time-sequence models to efficiently capture spatio-temporal relations between interactive agents and predict their dynamics. Our model infers dynamic interaction graphs in a latent space augmented with interpretable edge features that characterize the interactions. Moreover, we aim to enhance model interpretability and performance in OOD scenarios by disentangling the latent space of edge features, thereby strengthening model versatility and robustness. We validate our approach through extensive experiments on both simulated and real-world datasets. The results show superior performance compared to existing methods in modeling spatio-temporal relations, motion prediction, and identifying time-invariant latent features. 
\end{abstract}


\section{Introduction}\label{sec:intro}

Understanding and modeling complex interactions among dynamic agents is important to various applications and tasks, including robotics \citep{wang2022social}, traffic modeling and management \citep{binshaflout2023graph}, and social network analysis \citep{li2021relevance}. In the field of robotics, one essential downstream task is multi-agent trajectory prediction, which serves as a prerequisite for safe and high-quality decision-making and motion planning in complex and crowded scenarios. 
Modeling inter-agent interactions is crucial to understanding the joint dynamic behavior of the agents. For instance, the joint prediction of two vehicles approaching an intersection requires modeling and reasoning about their potential interactions, such as yielding, overtaking, or not interacting at all.

Recent approaches \citep{nri, dNRI, evolvegraph, sachdeva2022dider} have focused on modeling interactions among agents by inferring latent interaction graphs, where edges represent different types of interactions. However, these methods are limited to inferring categorical relations and are not equipped to capture more nuanced characteristics. For example, while these models can identify whether or not a pair of particles is connected by a spring, they are unable to infer its elastic coefficient. Furthermore, these methods focus on minimizing distance-based prediction errors in test cases that align with the observed data distribution. By primarily evaluating their methods using in-distribution samples, they overlook out-of-distribution situations, where models may encounter unseen and challenging interaction patterns between agents or diverse environmental contexts. This limits both the interpretability and generalizability of the existing works, which undermines the reliability of prediction models and proves inadequate for safety-critical applications such as autonomous driving.

\begin{figure}[!t]
    \centering
    \includegraphics[width=\columnwidth]{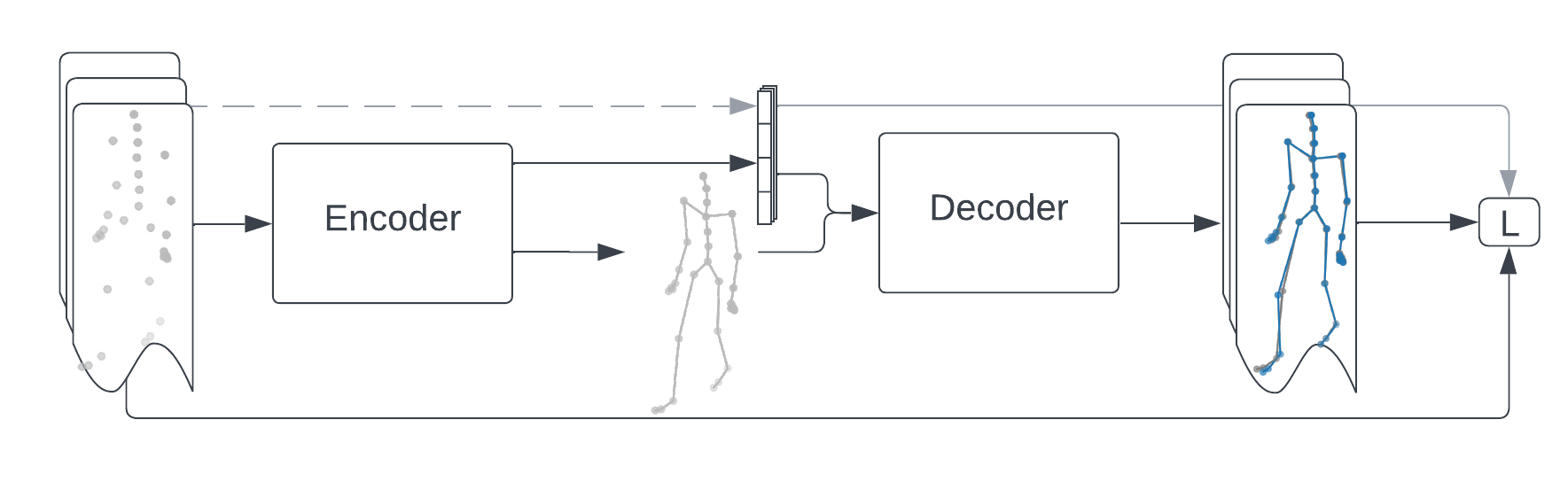}
    \caption{The encoder evaluates edge features, a section of which is used to increase interpretability through disentanglement, such as restricted labeling or pair matching.}
    \label{fig:teaser}  
\vspace{-0.2cm}
\end{figure}

In this work, we propose a model (dG-VAE) built upon a variational auto-encoder framework for discovering interpretable interactions from observations. Our model achieves interpretability by incorporating additional edge features that capture latent characteristics, such as the elastic coefficient of a spring. It combines interpretable graph structures and edge feature learning with multi-relational decoders within an unsupervised training framework. To enhance the interpretability of the learned edge features, we employ two types of disentanglement techniques, either supervised or unsupervised, depending on the application. By using an inference-based approach that disentangles time-invariant features, our model demonstrates interpretability and improved generalizability in out-of-distribution settings. Integrating interpretable components and inference-based learning allows our model to capture and understand more complex interactions, leading to more reliable and robust predictions.

The prospective applications of dG-VAE span various fields, opening up new avenues for exploration. In robotics, it has the potential to enhance safety by effectively forecasting the motion of other participants. In healthcare, it holds promise in refining gait analysis and contributing to the development of smart prosthetics and exoskeletons that sync with user intentions. In security, transportation, and traffic management, it could be used for identifying anomalies, strategizing crowd management, and optimizing traffic flows.

The main contributions of this paper are as follows:
\begin{itemize}
    \item We propose a VAE-based model architecture that goes beyond inferring dynamic interaction graphs by incorporating edge features to characterize the interactions.
    \item We leverage disentanglement techniques to partition the embedding space into time-invariant and temporal components, boosting prediction accuracy, enhancing out-of-distribution generalization, and occasionally yielding human-interpretable embeddings.
    \item We extensively evaluate our method on multiple benchmark datasets to show its effectiveness. Our approach outperforms baseline techniques in modeling spatio-temporal relations and accurately predicting interactions, trajectories, and underlying behavioral determinants.
\end{itemize}

\section{Related work}\label{sec:related_work}

\emph{Trajectory prediction} is important in fields like robotics, autonomous vehicles, and human interaction. Social LSTM \citep{alahi2016} uses long short-term memory networks to model pedestrian interactions and predict their trajectories in crowded spaces. Building upon this framework, Trajectron \citep{ivanovic19} uses dynamic spatio-temporal graphs and graph-structured LSTM networks to model and predict the trajectories of multiple agents in a scene. 
Generative adversarial networks have been used to predict trajectories compliant with social and physical constraints \citep{gupta18, sadeghian2019}. Recent works proposed more advanced graph- or transformer-based approaches to model the spatio-temporal relations between interactive agents, thereby enhancing prediction performance \citep{mohamed2020social, evolvegraph, yu2020spatio, li2021rain, zhou2022grouptron, imma}. 

\emph{Spatio-temporal graph modeling} has been widely studied to capture the spatio-temporal relations between dynamic interacting agents. Social-STGCNN \citep{mohamed2020social} introduced a spatio-temporal graph convolutional neural network for human trajectory prediction. EvolveGraph \citep{evolvegraph} and dNRI \citep{dNRI} are dynamic relational reasoning methods to identify underlying relations based on the information encoded in given sets of trajectories. A spatio-temporal attention mechanism \citep{li2021spatio} and a spatio-temporal graph transformer network \citep{yu2020spatio} were proposed to model the inter-graph temporal dependencies.

The significance of \emph{interpretability} is increasingly recognized in various fields such as autonomous vehicles~\citep{sachdeva2023rank2tell} or security systems~\citep{samek2017explainable} for example. Specifically, in motion prediction, this emphasis on interpretability has given rise to methods like GRIT \citep{grit}, which use decision trees to better understand learned spaces. \citet{parth21} combine rule-based and neural network models to predict high-level intents and scene-specific residuals. For multi-agent interaction modeling, studies such as Grounded Relational Inference \citep{tang21}, which learns reward functions, and GRIN \citep{grin}, disentangling interactions from agent intentions, are relatively recent. Collaborative Prediction Units \citep{li_cpu}, learning a weighted aggregate of individual prediction units, consider inter-agent influences and adjust in real-time to incoming data. Other works \citep{lingfeng22, lee19} annotate ground truth or pseudo-labels to interactions for better understanding. Despite these advancements, most methods view interactions as static, categorical, or both. In contrast, our research uncovers disentangled and interpretable sequences of dynamic interactions from multi-agent observations.

Our work lies at the intersection of these three fields. We build on DESIRE \citep{desire}, which leverages a conditional VAE to forecast future trajectories amid multiple interacting agents, an approach later enhanced by NRI \citep{nri} and dNRI \citep{dNRI} through the integration of Graph Neural Networks (GNNs) to deduce a latent graph. These works rely on the proficiency of VAEs to comprehend intricate data distributions and produce new samples resembling original data. Our work builds on these models, enhancing the predicted latent graph with edge features, thereby boosting expressivity, and introducing ``disentanglement'' of the embedding space to promote the discovery of interpretable factorized latent representations. Disentanglement, initially an adjustable hyperparameter in the loss function, has shown great success in computer vision \citep{higgins2017betavae} and reinforcement learning \citep{higgins2017darla}.

\section{Preliminaries}\label{sec:preliminaries}

\emph{Variational auto-encoders} (VAEs) are a type of deep generative model that blends the concepts from deep learning and probabilistic graphical models. VAEs consist of an encoder and a decoder. The {encoder} maps the input data $x$ to a lower-dimensional latent variable $z$, represented by a probability distribution $q_\theta(z\mid x)$, typically a Gaussian distribution with a learnable mean and covariance. It learns to approximate the true posterior distribution of the latent variable given the input data. The {decoder} maps the latent variable back to the original data space, i.e., it learns the conditional probability distribution $ p_\phi(x\mid z)$ of the input data given the latent variable.

During training, we optimize the \textit{reconstruction} objective, which ensures that the generated samples resemble the original data, and the \textit{regularization} objective, which encourages the learned latent space to have a specific structure $p(z)$ (e.g., a standard Gaussian distribution), simultaneously. The resulting loss function is written as
\begin{equation}
    l_i(\theta,\phi) = -\E_{z\sim q_\theta(z\mid x)}\big[\log p_\phi(x\mid z)\big] + \textsc{KL}\big(q_\theta(z\mid x)\Vert p(z)\big),\label{eq:elbo_def}
\end{equation}
a combination of the reconstruction error and the Kullback-Leibler (KL) divergence between the approximate and true posterior distributions, which is also known as the evidence lower bound (ELBO).

\emph{Graph neural networks} (GNNs) are a class of deep learning models designed to handle data represented as graphs. They are particularly well-suited for problems where the data has a complex, irregular structure. GNNs process graph-structured data by iteratively passing and aggregating messages between neighboring nodes.

Let $V$ denote the set of nodes in the graph. During each message-passing iteration, a hidden embedding $\mathbf{x}_u$ corresponding to node $u \in V$ is updated according to information aggregated from $u$’s neighborhood nodes $\mathcal{N}(u)$. The message-passing update can be expressed as
\begin{equation}\label{eq:gnn_def}
    \mathbf{x}_u^{(k+1)} = \textsc{up}^{(k)}\left(\mathbf{x}_u^{(k)}, \textsc{agg}^{(k)}\left(\{\mathbf{x}_v^{(k)} \mid v \in \mathcal{N}(u)\}\right)\right),
\end{equation}
where \textsc{up}, short for \textsc{update}, and \textsc{agg} denote arbitrary differentiable functions (e.g., MLP), and $k$ denotes the index of message-passing layers. This mechanism can be applied iteratively to capture information from a broader neighborhood, and the final node representations or aggregated graph representations can be used for various tasks, such as node classification, link prediction, and graph classification.

\section{Method}\label{sec:method} 

\textbf{Problem Definition.} Our goal is to learn accurate trajectory distributions for a variable number of interacting agents $N(t)$ based on historical observations. 
For each agent $i$ present at time step $t$, we consider its history $x^{(i)}_{t-H+1:t}$ where $H$ denotes the history horizon. 
We aim to predict the distribution of all entities' future states  $p(y^{(i)} \mid \{x^{(j)}_{t-H+1:t} \mid \forall j \in \{1,\ldots, N(t)\}\})$ for the upcoming $T$ steps. Here $y^{(i)} =\hat{x}^{(i)}_{t+1:t+T} \in R^{T\times N(t)\times d}$, with $d$ the state dimension, represents the predicted future trajectory of agent $i$.

\textbf{Overview.} 
Our proposed model architecture is inspired by NRI \citep{nri} and dNRI \citep{dNRI}. While these previous iterations primarily focused on edge classification, our method uniquely emphasizes learning edge features, which can be leveraged further. This modification increases the expressivity of the network and allows for the incorporation of disentanglement in the edge features to specific characteristics of the underlying interactions. We hypothesize that this added regularized expressivity increases the out-of-distribution generalization ability and interpretability of the learned embeddings. 

We provide the model with historical input $X\in\R^{N \times H\times d}$. The input is passed through the encoder, which computes edge logits $p({z}_t\mid X_{t-H+1:t})$, i.e., the likelihood of this edge existing in the latent space, alongside an edge feature matrix $E_t$. We then sample $z_t$ from the posterior distribution and pass ${z_t}$ and $E_{z, t}$ to the decoder, which predicts the next state $X' \in \R^{N \times d}$. This process is shown in \cref{fig:teaser}.

\textbf{Encoder.}
The architecture of the encoder builds upon dNRI's method. We use a GNN block composed of a linear embedding followed by an edge convolutional layer \citep{edgeconv}:
\begin{equation}
    \mathbf{x}^{\prime}_{i, t} = \sum_{j \in \mathcal{N}_t(i)} h_{\mathbf{\Theta}}\left(\mathbf{x}_{i, t} \, \Vert \, \mathbf{x}_{j, t} - \mathbf{x}_{i, t}\right),
\end{equation}
where $h_{\mathbf{\Theta}}$ denotes a neural network (i.e., an MLP) and $\Vert$ indicates the concatenation of two node embeddings $\mathbf{x}_{i, t}$ and $\mathbf{x}_{j, t}$. 
We recall from \cref{sec:preliminaries} that GNNs aggregate messages from neighboring nodes to update the node embeddings at each iteration. Because we are learning edge features instead of node features, we adapt the last \textit{edge convolutional} layer to edge-level embeddings, which is written as
\begin{equation}
    \mathbf{e}^{\prime}_{ij, t} = h_{\mathbf{\Theta}}\left(\mathbf{x}_{i, t} \, \Vert \, \mathbf{x}_{j, t}\right).
\end{equation}

The intermediate embedding we obtain from the GNN block is then forwarded to an RNN block composed of a single LSTM layer with ELU activation and a dropout layer. Next, we store the RNN hidden state and give the embeddings to two MLP heads each composed of two linear layers with ELU activations. The first returns the edge posterior logits and the second computes the edge features $E_{t}$.

\begin{figure}[t]
    \centering
    \includegraphics[width=\columnwidth]{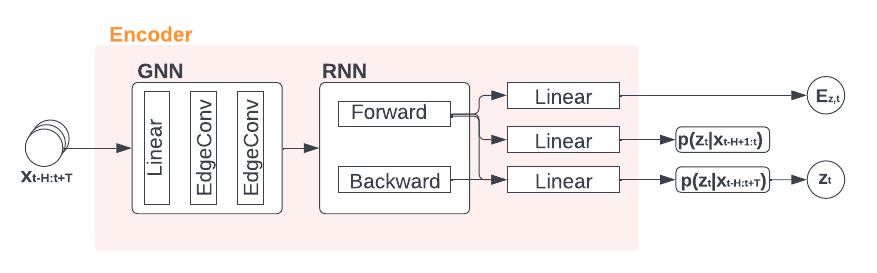}
    \caption{Encoder architecture that learns the prior.}
    \label{fig:encoder_alt}
\vspace{-0.2cm}
\end{figure}

The dNRI method \citep{dNRI} suggested \emph{learning the prior}, an idea we incorporated into our encoder shown in \cref{fig:encoder_alt}. This alteration requires a forward and a {backward RNN} to generate two embeddings. The forward embedding is used to a) generate a prior, and b) in conjunction with the backward embedding, return the posterior $p({z}_t\mid X_{t-H+1:t})$. As a result of using the backward RNN, the posterior is computed using future information. During training, the posterior is used the way it is currently in our model, and the prior is trained to match it. During testing, the latent variable $z_t$ is sampled from the learned prior.

\textbf{Sampling.}
The encoder returns a distribution over possible relations. Because traditional categorical sampling is not differentiable, required for backpropagating weight updates, we adopt the concrete distribution \citep{maddison2016concrete}, a continuous proxy to the discrete categorical distribution. This sampling technique uses reparameterization by, first, sampling a vector $\mathbf{g}$ from $\textsc{gumbel}(0, 1)$ and then calculating
\begin{equation}
\mathbf{z}{ij, t} = \textsc{softmax}\left((\mathbf{\hat{z}}{ij, t} + \mathbf{g})/\tau\right),
\end{equation}
where $\mathbf{\hat{z}}^{(ij)}_t$ are the posterior logits at time $t$ and $\tau$ adjusts distribution smoothness. This method approximates discrete sampling in a gradient-friendly manner, allowing the encoder to receive feedback from decoder reconstruction.

\textbf{Decoder.}
The decoder takes as inputs graph $z_t$, node features $X_t$, and edge features $E_{z, t}$. These are processed by an adapted \textit{edge convolutional} layer: 
\begin{equation}
    \mathbf{x}^{\prime}_{i, t} = \sum_{j \in \mathcal{N}_t(i) \mid  \mathbf{z}_t} h_{\mathbf{\Theta}}\left(\mathbf{x}_{i, t} \, \Vert \, \mathbf{x}_{j, t} \, \Vert \, \mathbf{e}_{ij, t}\right),
\end{equation}
to obtain node-level embeddings, which are then forwarded to a GRU layer with ELU activation and a linear readout.

\textbf{Loss function and training strategy.}
The loss function is composed of a generative loss, which measures the difference between the model's input and output, and a latent loss, which compares the latent vector to a Gaussian distribution with zero mean and unit variance.
We use the negative log-likelihood loss:
\begin{equation}
    \textsc{NLL} = \frac{1}{T} \sum_{t\in[T]} \sum_{j\in [d]} \frac{1}{2\sigma}({z}_{t,j} - {y}_{t,j})^2,
\end{equation}
which is widely used when training probabilistic models. Here, $\sigma = \num[round-precision=1]{5e-5}$ is a hyperparameter. We used the learned prior when evaluating the KL-divergence (i.e., the latent loss) of $p(z\mid X)$ and $p(z)$:
\begin{equation}
    \textsc{KL} = \sum_{t\in[T]} \left(-\sum_{{z}} p({z}_t\mid X)\log p({z}_t\mid X, {z}_{1:t-1})\right).
\end{equation}

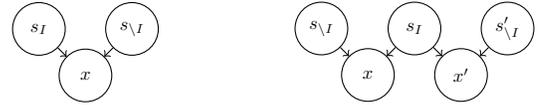
\begin{figure}[t]
\centering
    \begin{subfigure}[b]{0.48\linewidth}
        \centering
        \begin{tikzpicture}[main/.style = {draw, circle, scale=0.7, minimum width=1cm,node distance=1.25cm and 1.25cm}]
        \node[main] (1) {$x$}; 
        \node[main] [above right of=1] (2) {$s_{\backslash I}$}; 
        \node[main] [above left of=1] (3) {$s_I$}; 
        \draw[->] (2) -- (1);
        \draw[->] (3) -- (1);
        \end{tikzpicture} 
        \caption{Restricted Labeling $p(x, s_I)$}
        \label{fig:restricted_labeling}
    \end{subfigure}
    \begin{subfigure}[b]{0.48\linewidth}
        \centering
        \begin{tikzpicture}[main/.style = {draw, circle, scale=0.7, minimum width=1cm,node distance=1.25cm and 1.25cm}]
        \node[main] (1) {$x$}; 
        \node[main] [above left of=1] (3) {$s_{\backslash I}$}; 
        \node[main] [above right of=1] (4) {$s_I$}; 
        \node[main] [below right of=4] (2) {$x'$}; 
        \node[main] [above right of=2] (5) {$s'_{\backslash I}$};
        \draw[->] (3) -- (1);
        \draw[->] (4) -- (1);
        \draw[->] (4) -- (2);
        \draw[->] (5) -- (2);
        \end{tikzpicture} 
        \caption{Pair Matching $p(x, x')$}
        \label{fig:pair_matching}
    \end{subfigure}
    \caption{Variations of disentanglement.}
    \vspace{-0.2cm}
    \label{fig:disentanglement}
\end{figure}

\textbf{Disentanglement.} 
To encourage interpretable edge feature learning, we introduce two types of disentanglement: \textit{restricted labeling} and \textit{pair matching}, as shown in \cref{fig:disentanglement}. The former, a supervised method from style-content disentanglement \citep{gabbay2020}, matches distributions based on observed $x$ and a subset of ground truth features $s_I$. As the encoder processes the edge feature matrix $E$ from \cref{fig:encoder_alt}, certain learned variables are directed to align with specified attributes using an additional loss. For example, in the Spring dataset (\cref{ssec:spring}), we apply regression to parts of the predicted feature matrix $E$ guided by the known spring constants during training. Although the actual spring constants are unknown during testing, the model has been conditioned to infer them. By basing the model on a feature with known semantics and predictive implications, we can further enhance the interpretability and performance of our model.

\textit{Pair Matching} is an unsupervised approach, focusing on paired data $(x,x')$ with a subset of common feature values. It can be deemed a weakly supervised approach as it does not rely on the underlying values of $s_I$ but correlates with its indices. Although various strategies exist for its implementation, this method typically involves sampling twice from the encoder and equating dimensions between $E_1$ and $E_2 \in \R^{m\times T\times d}$, where $m$ is the number of edges and $d$ is the hidden dimension, by averaging their differences. Given our model's $T$ sequential samples, we forgo double sampling. We average the partial matrix $E_h \in \R^{m\times T\times d_h}$ across time, resulting in $\hat{E}_h \in \R^{m\times d_h}$, and populate $E$ with these time-invariant features, whereby the model learns time-invariant latent characteristics of each interaction. Our strategy thereby renders dimensions $[d_h] \subset [d]$ virtually static.

\section{Experiments}\label{sec:results}

\textbf{Datasets and Baselines.}
We evaluate our method on four benchmark datasets: NBA, Spring, Motion Capture, and inD datasets through comparisons with the following baselines: a) A two-layer MLP with a hidden dimension of 256 and ReLU activation, b) an LSTM layer, with a hidden dimension of 128 between two fully connected layers with hidden dimensions of 64, that processes all agents simultaneously, c) IMMA \citep{imma}, a forward prediction model that uses a multiplex latent graph to represent different types of interactions and attention, and d) dNRI \citep{dNRI}, a VAE model with recurrent GNN modules that re-evaluates the predicted latent graph dynamically.

The first two are deterministic, non-variational baselines, while IMMA and dNRI are generative models. We further evaluate our method with and without disentanglement. As restricted labeling requires ground truth labels to be provided for each interaction, it is reserved for certain datasets, such as the Spring dataset in \cref{ssec:spring}. Otherwise, we resort to pair matching, an unsupervised approach, which we use to learn a set of time-invariant variables of the feature space that characterize specific aspects of certain interactions.

\textbf{Ablation Studies.} dVAE is a variant of dG-VAE, characterized by its lack of edge features. When referring to dG-VAE, unless otherwise specified, we are addressing the version without disentanglement. Therefore, ablation studies are intrinsically incorporated and have not been neglected.

\textbf{Evaluation.}
As in prior work \citep{evolvegraph, ivanovic19}, we evaluate the standard metrics used in trajectory forecasting: 1) The minimum average displacement error (minADE), which refers to the mean Euclidean distance between the ground truth and predicted trajectories, 2) the minimum final displacement error (minFDE), which refers to the Euclidean distance between the predicted final position and the ground truth at the prediction horizon, and 3) the root mean squared error (RMSE).  We also evaluate graph accuracy, the percentage overlap between edges present in the ground truth graph and those present in the inferred graph, for the Spring dataset only, as it is the only one that has ground truth graphs. For minADE and minFDE, we evaluate $k=20$ randomly sampled predicted trajectories and choose the best value in the set. We report the mean and standard error across 20 test subsets for each metric.

\textbf{Implementation.}
dG-VAE consists of a GNN component with a hidden dimension of 128, a RNN with a hidden dimension of 64, and read-out heads with a hidden dimension of 256. The dimension of edge feature embedding is set to 16 for the Spring dataset, 32 for the NBA dataset, 32 for the Motion Capture dataset, and 64 for the inD dataset, respectively. We allocated 16 nodes to temporal pair matching, when applicable.  The decoder RNN has a hidden dimension of 256.  The sampler uses a Gumbel temperature of $0.5$, which is the same as dNRI. The batch size and learning rate are set to 128, $2\times 10^{-4}$ for NBA, Spring, and Motion Capture datasets, and 1, $5\times 10^{-4}$ for the inD dataset.
While training multiple edge types is possible, we only evaluate binary edge prediction to evaluate the impact of learned edge features without correlation bias. We trained each model with the Adam optimizer for 150 epochs for NBA, Spring, and Motion Capture, and 400 epochs for inD. We used a GeForce RTX 3080 to train and evaluate all experiments.

\subsection{NBA}\label{ssec:nba}
The NBA dataset contains 100K examples with an 80/10/10 train/validation/test split. It features the trajectories of all ten players and the ball. Each trajectory has $50$ time steps at a frequency of $\qty[round-precision=1]{3}{\Hz}$, which translates to a prediction horizon of $\qty[round-precision=1]{3.6}{\second}$. We normalize and mean-shift the position and velocity to the range of $[-1, 1]$.  The preprocessed data can be downloaded \hyperlink{https://drive.google.com/drive/folders/1OFEle-qjXus_FtKvmDSh1qdqfnk1HtPB?usp=sharing}{here}. Unlike some other works that used this dataset, our study is not limited to the offensive team's coordination on a half-court, ignoring defensive players and the ball, but analyzes the performance of semantically understanding the whole move.

\begin{figure}[t!]
    \hspace*{-0.0cm}\begin{subfigure}[b]{0.3\linewidth}
        \centering
        \begin{tikzpicture}[rotate=90]
            \begin{axis} [ width=6.5cm, axis equal,
                        hide x axis, hide y axis,
                        ]
        	    \addplot graphics[xmin=-5,ymin=-5,xmax=89,ymax=55] {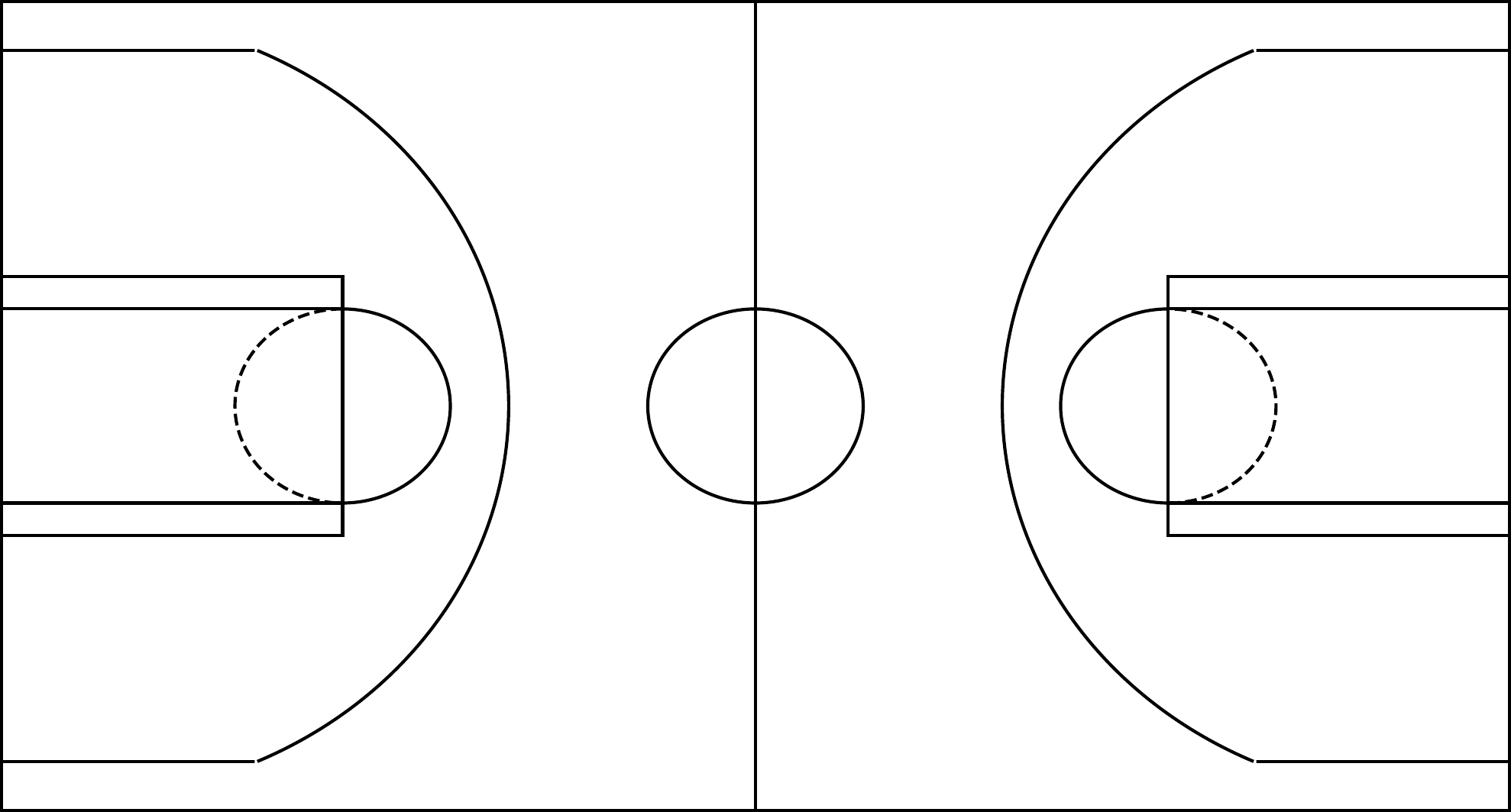};
                \pgfplotstableread[col sep=comma, header=true]{ieeeconf/data/dnri_nba.csv}\datatable
               \foreach \i in {0}{
                    \pgfmathtruncatemacro{\xxcol}{\i*2+11}
                    \pgfmathtruncatemacro{\yycol}{\i*2+12}
                    \addplot+ [pastelMagenta,solid, very thick,opacity=0.5,no marks] table [x index=\xxcol, y index=\yycol] {\datatable};
               }
               \foreach \i in {1,...,5}{
                    \pgfmathtruncatemacro{\xxcol}{\i*2+11}
                    \pgfmathtruncatemacro{\yycol}{\i*2+12}
                    \addplot+ [pastelGreen, solid,very thick,opacity=0.5,no marks] table [x index=\xxcol, y index=\yycol] {\datatable};
               }
               \foreach \i in {6,...,10}{
                    \pgfmathtruncatemacro{\xxcol}{\i*2+11}
                    \pgfmathtruncatemacro{\yycol}{\i*2+12}
                    \addplot+ [pastelBlue, solid,very thick,opacity=0.5,no marks] table [x index=\xxcol, y index=\yycol] {\datatable};
               }
                \foreach \i in {0,...,10}{
                	\pgfmathtruncatemacro{\xcol}{\i*2+1}
                	\pgfmathtruncatemacro{\ycol}{\i*2+2}
                	\addplot+ [gray,dashed,very thick,opacity=0.5,no marks] table [x index=\xcol, y index=\ycol] {\datatable};
                }
            \end{axis}
        \end{tikzpicture}
        \caption{dNRI}
        \label{fig:nba_dNRI}
    \end{subfigure}
    \hfill
    \hspace*{-0.0cm}\begin{subfigure}[b]{0.3\linewidth}
        \centering
        \begin{tikzpicture}[rotate=90]
            \begin{axis} [ width=6.5cm, axis equal,
                        hide x axis, hide y axis,
                        ]
        	    \addplot graphics[xmin=-5,ymin=-5,xmax=89,ymax=55] {ieeeconf/figs/fullcourt.pdf};
                \pgfplotstableread[col sep=comma, header=true]{ieeeconf/data/dvae_pm_nba.csv}\datatable
               \foreach \i in {0}{
                    \pgfmathtruncatemacro{\xxcol}{\i*2+11}
                    \pgfmathtruncatemacro{\yycol}{\i*2+12}
                    \addplot+ [pastelMagenta,solid, very thick,opacity=0.5,no marks] table [x index=\xxcol, y index=\yycol] {\datatable};
               }
               \foreach \i in {1,...,5}{
                    \pgfmathtruncatemacro{\xxcol}{\i*2+11}
                    \pgfmathtruncatemacro{\yycol}{\i*2+12}
                    \addplot+ [pastelGreen,solid, very thick,opacity=0.5,no marks] table [x index=\xxcol, y index=\yycol] {\datatable};
               }
               \foreach \i in {6,...,10}{
                    \pgfmathtruncatemacro{\xxcol}{\i*2+11}
                    \pgfmathtruncatemacro{\yycol}{\i*2+12}
                    \addplot+ [pastelBlue, solid,very thick,opacity=0.5,no marks] table [x index=\xxcol, y index=\yycol] {\datatable};
               }
                \foreach \i in {0,...,10}{
                	\pgfmathtruncatemacro{\xcol}{\i*2+1}
                	\pgfmathtruncatemacro{\ycol}{\i*2+2}
                	\addplot+ [gray,dashed,very thick,opacity=0.5,no marks] table [x index=\xcol, y index=\ycol] {\datatable};
                }
            \end{axis}
        \end{tikzpicture}
        \caption{\ours}
        \label{fig:nba_dVAE}
    \end{subfigure}
    \hfill
    \hspace*{-0.0cm}\begin{subfigure}[b]{0.3\linewidth}
        \centering\begin{tikzpicture}[rotate=90]
            \begin{axis} [ width=6.5cm, axis equal,
                        hide x axis, hide y axis,
                        ]
        	    \addplot graphics[xmin=-5,ymin=-5,xmax=89,ymax=55] {ieeeconf/figs/fullcourt.pdf};
                \pgfplotstableread[col sep=comma, header=true]{ieeeconf/data/lstm_nba.csv}\datatable
               \foreach \i in {0}{
                    \pgfmathtruncatemacro{\xxcol}{\i*2+11}
                    \pgfmathtruncatemacro{\yycol}{\i*2+12}
                    \addplot+ [pastelMagenta, solid, very thick, opacity=0.5, no marks] table [x index=\xxcol, y index=\yycol] {\datatable};
               }
               \foreach \i in {1,...,5}{
                    \pgfmathtruncatemacro{\xxcol}{\i*2+11}
                    \pgfmathtruncatemacro{\yycol}{\i*2+12}
                    \addplot+ [pastelGreen, solid, very thick, opacity=0.5, no marks] table [x index=\xxcol, y index=\yycol] {\datatable};
               }
               \foreach \i in {6,...,10}{
                    \pgfmathtruncatemacro{\xxcol}{\i*2+11}
                    \pgfmathtruncatemacro{\yycol}{\i*2+12}
                    \addplot+ [pastelBlue, solid, very thick, opacity=0.5, no marks] table [x index=\xxcol, y index=\yycol] {\datatable};
               }
                \foreach \i in {0,...,10}{
                	\pgfmathtruncatemacro{\xxcol}{\i*2+1}
                	\pgfmathtruncatemacro{\yycol}{\i*2+2}
                	\addplot+ [gray, dashed, very thick, opacity=0.5, no marks] table [x index=\xxcol, y index=\yycol] {\datatable};
                }
            \end{axis}
        \end{tikzpicture}
        \caption{LSTM}
        \label{fig:nba_lstm}
    \end{subfigure}
    \caption{(NBA) Trajectory samples predicted by different models. The grey lines represent ground truth trajectories and the blue and green lines show the predictions for home and visiting teams. The purple lines represent the basketball.}
    \label{fig:nba_performance_samples}
\vspace{-0.4cm}
\end{figure}
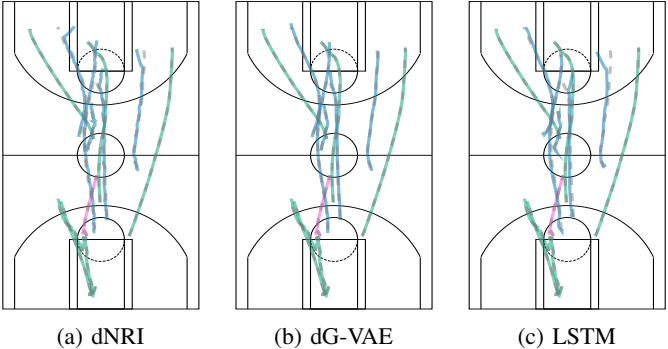


\begin{table*}[ht]
  \caption{Performance comparison for NBA dataset}
  \label{tab:nba_performance_comparison}
  \centering
  \begin{tabular}{@{}p{35mm}|S[table-column-width=7ex, round-precision=3]@{\,}c @{\,}S[round-precision=2, table-column-width=9ex]S[table-column-width=7ex, round-precision=3]@{\,}c @{\,}S[round-precision=2, table-column-width=9ex]S[table-column-width=7ex, round-precision=3]@{\,}c @{\,}S[round-precision=2, table-column-width=9ex]S@{}}
    \toprule
    {Model} & \multicolumn{3}{c}{RMSE [m]} &  \multicolumn{3}{c}{min ADE [m]} & \multicolumn{3}{c}{min FDE [m]} & {Connectivity} \\ 
    \midrule 
     {Linear} & 1.0270254850387573 &$\pm$ & \num{1.0013713229919634e-2} & 1.0376331806 &$\pm$ & \num{9.44517057798506e-3} & 4.1184620857 &$\pm$ & \num{1.5168391150962501e-2} & {--}\\ 
    {LSTM} & 0.8841372728347778 &$\pm$ & \num{3.5471823908444053e-3} & 0.8303674459 &$\pm$ & \num{2.0157312437877e-3} & 2.8948991299 &$\pm$ & \num{1.0309563259144082e-2}&{--}\\ 
    {IMMA} &  1.035659909248352 &$\pm$ & \num{9.412177937704026e-3}  & 0.9654775262 &$\pm$ & \num{9.992779101030734e-3}& 2.9215687513 &$\pm$ & \num{1.745871696524536e-2}& {--}\\
    {dNRI} & 1.1404260396957397 &$\pm$ & \num{5.086078685745477e-3}& 0.8016893268 &$\pm$ & \num{3.1840740484497766e-3} & 4.0801806450 &$\pm$ & \num{2.050223992540724e-2}& 0.6784\\
    dG-VAE (ours) & 0.8764407634735107 &$\pm$ & \num{4.191203194011451e-03} & 0.5467479229 & $\pm$ & \num{1.837328000101761e-3} & 3.1986761093 &$\pm$ & \num{1.3880909320785877e-2}& 1.000 \\
    dG-VAE + pair matching & \B 0.5790384411811829 &$\pm$ & \num{4.061595021551868e-3} &  \B 0.3814821541 &$\pm$ & \num{1.7756670320061483e-3} & \B 2.1661968231 &$\pm$ & \num{1.4758361779618695e-2} & 0.9989\\
    \bottomrule
  \end{tabular}
\end{table*}

\cref{fig:nba_performance_samples} shows trajectories generated by each model, highlighting notable differences. It is evident that our approach demonstrates a smaller deviation from the ground truth trajectories compared to dNRI and LSTM. 
\cref{tab:nba_performance_comparison} summarizes the {testing} statistics for different models where the best values for each metric are highlighted in bold. We note that our method outperforms the strongest baseline (dNRI) in all metrics by around 25\% when edge features are learned and around 45\% when these features are disentangled into temporal and static features, i.e., pair matched.

Further analysis showed that dNRI, which only predicts binary edges converges to a 67\% graph connectivity, while dG-VAE convergences to a fully connected graph. This implies that all agents are to be considered when predicting a player's next moves. Our method allows for this ``all interactions are relevant'' state, as the augmented edge features add the expressivity necessary to distinguish between interactions. 

\subsection{k-Spring}\label{ssec:spring}

The Spring dataset is composed of 70K total rollouts of simulated systems with $N$ particles, with 50K for training and 10K for testing and validation each. In our experiments, we use $N=5$. No external forces are applied, except elastic collisions with the box boundaries. 
With a probability of $0.7$, we connect each pair of particles with a spring. The interaction between particles linked by springs is governed by Hooke's law, i.e., $F_{ij} = -k \cdot (r_i - r_j)$. Here, $F_{ij}$ represents the force exerted by particle $v_j$ on particle $v_i$. The spring constant is denoted by $k$ and is uniformly sampled between 0.5 and 2, and $r_i$ indicates the 2D coordinate of particle $v_i$. The out-of-distribution data is sampled with a 0.5 probability of connection and spring constants of 1, 2, or 3.
Given the initial locations and velocities, which are sampled from a multi-variate Gaussian, we simulate $50$ time steps at a frequency of $\qty[round-precision=1]{6}{\Hz}$. We provide the preprocessed data \hyperlink{https://drive.google.com/drive/folders/1IYdUd4QETpTw7uu1nqI8AO_hfKzKTJah?usp=sharing}{here}.

\begin{figure}[!tbp]
    \hspace*{-0.0cm}\begin{subfigure}[b]{0.32\linewidth}
        \centering
        \begin{tikzpicture}[scale=0.9]
        \begin{axis}[width=4.5cm, height=6cm,
                    yticklabels={}, xticklabels={}, 
                    axis equal,
                    cycle list name=pastelcolors    
                    ]
            \pgfplotstableread[col sep=comma,header=true]{ieeeconf/data/dnri_springs5.csv}\datatable
            \foreach \i in {0,...,4}{
                \pgfmathtruncatemacro{\xxcol}{\i*2+11}
                \pgfmathtruncatemacro{\yycol}{\i*2+12}
                \addplot+ [very thick,opacity=0.5,no marks] table [x index=\xxcol, y index=\yycol] {\datatable};
            }
            \foreach \i in {0,...,4}{
                \pgfmathtruncatemacro{\xcol}{\i*2+1}
                \pgfmathtruncatemacro{\ycol}{\i*2+2}
                \addplot+ [gray,dashed,very thick,opacity=0.5,no marks] table [x index=\xcol, y index=\ycol] {\datatable};
            }
        \end{axis}
        \end{tikzpicture}
        \caption{dNRI}
        \label{fig:spring_dNRI}
    \end{subfigure}
    \hfill
    \hspace*{-0.0cm}\begin{subfigure}[b]{0.32\linewidth}
        \centering\begin{tikzpicture}[scale=0.9]
        \begin{axis}[width=4.5cm, height=6cm,
                    yticklabels={}, xticklabels={}, 
                    axis equal,
                    cycle list name=pastelcolors 
                    ]

            \pgfplotstableread[col sep=comma,header=true]{ieeeconf/data/svae_springs5.csv}\datatable
            \foreach \i in {0,...,4}{
                \pgfmathtruncatemacro{\xxcol}{\i*2+11}
                \pgfmathtruncatemacro{\yycol}{\i*2+12}
                \addplot+ [very thick,opacity=0.5,no marks] table [x index=\xxcol, y index=\yycol] {\datatable};
            }
            \foreach \i in {0,...,4}{
                \pgfmathtruncatemacro{\xcol}{\i*2+1}
                \pgfmathtruncatemacro{\ycol}{\i*2+2}
                \addplot+ [gray,dashed,very thick,opacity=0.5,no marks] table [x index=\xcol, y index=\ycol] {\datatable};
            }
        \end{axis}
        \end{tikzpicture}
        \caption{\ours}
        \label{fig:spring_svae}
    \end{subfigure}
    \hfill
    \hspace*{-0cm}\begin{subfigure}[b]{0.32\linewidth}
        \centering
        \begin{tikzpicture}[scale=0.9]
        \begin{axis}[width=4.5cm, height=6cm,
                    yticklabels={}, xticklabels={}, 
                    axis equal,
                    cycle list name=pastelcolors 
                    ]
            \pgfplotstableread[col sep=comma,header=true]{ieeeconf/data/lstm_springs5.csv}\datatable
            \foreach \i in {0,...,4}{
                \pgfmathtruncatemacro{\xxcol}{\i*2+11}
                \pgfmathtruncatemacro{\yycol}{\i*2+12}
                \addplot+ [very thick,opacity=0.5,no marks] table [x index=\xxcol, y index=\yycol] {\datatable};
            }
            \foreach \i in {0,...,4}{
                \pgfmathtruncatemacro{\xcol}{\i*2+1}
                \pgfmathtruncatemacro{\ycol}{\i*2+2}
                \addplot+ [gray,dashed,very thick,opacity=0.5,no marks] table [x index=\xcol, y index=\ycol] {\datatable};
            }
        \end{axis}
        \end{tikzpicture}
        \caption{LSTM}
        \label{fig:spring_lstm}
    \end{subfigure}
    \caption{($k$-Spring) Trajectory samples predicted by different models. Each color represents a different point mass and the grey lines represent the ground truth trajectories.} 
    \label{fig:spring_performance_samples}
\end{figure}
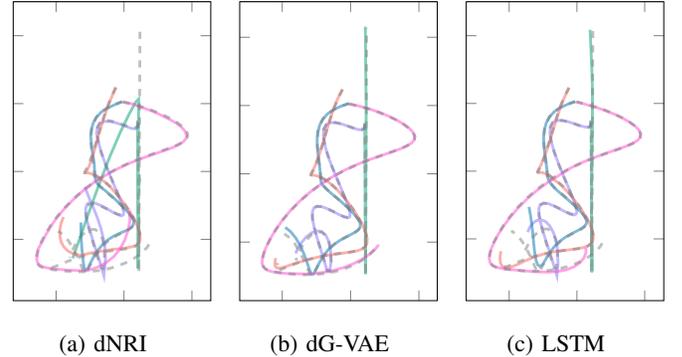

\begin{table*}[h]
  \caption{Performance comparison for 5-Spring dataset}
  \label{tab:spring_performance_comparison}
  \centering
  \sisetup{separate-uncertainty,table-align-uncertainty}
  \begin{tabular}{@{}p{30mm}|S[table-column-width=8ex, round-precision=3]@{\,} c @{\,}S[round-precision=2, table-column-width=9ex]S[table-column-width=8ex, round-precision=3]@{\,} c @{\,}S[round-precision=2, table-column-width=9ex]S[table-column-width=8ex, round-precision=3]@{\,} c @{\,}S[round-precision=2, table-column-width=10ex]|S[table-column-width=8ex, round-precision=3]@{\,} c @{\,}S[round-precision=2, table-column-width=10ex]S[table-column-width=8ex, round-precision=3]@{\,} c @{\,}S[round-precision=2, table-column-width=10ex]@{}}
    \toprule
    & \multicolumn{10}{c}{} & \multicolumn{3}{c}{{OOD}}\\
    {Model} & \multicolumn{3}{c}{RMSE [m]} & \multicolumn{3}{c}{min ADE [m]} & \multicolumn{3}{c|}{min FDE [m]}  & \multicolumn{3}{c}{{min ADE [m]}} & \multicolumn{3}{c}{{min FDE [m]}}\\
    \midrule
    {Linear} & 0.19561775028705597 &$\pm$& \num{9.357313418520296e-4} & 0.2017308027 &$\pm$& \num{9.807627097696326e-4} & 0.4456377625  &$\pm$& \num{2.3053591777618338e-3} &  2.1575637 &$\pm$& \num{1.0295829584280713e-2} & 3.5406818 &$\pm$& \num{2.0183227491667066e-2} \\ 
    {LSTM} & 0.03655025735497475 &$\pm$& \num{5.11963984868644e-3} & 0.0243643243 &$\pm$& \num{6.5367265193217085e-3} & 0.1593434811 &$\pm$& \num{1.3334506149079276e-2} & 0.40582213 &$\pm$& \num{1.3856837071764205e-3} & 1.4847838 &$\pm$& \num{1.1280437527257034e-2} \\ 
    {IMMA} &  0.16189533472061157 &$\pm$& \num{1.1481372386031256e-3} &  0.1756591201 &$\pm$& \num{1.2465653792915722e-3} & 0.3495028615 &$\pm$& \num{2.660744566733948e-3} & 1.3891094 &$\pm$& \num{5.74727137224308e-3} & 1.9340849 &$\pm$& \num{1.1614918775348685e-2}\\ 
    {dNRI} & 0.11395442485809326 &$\pm$& \num{4.6935335728201023e-4} & 0.0577938184 &$\pm$& \num{2.3169115373865843e-4} & 0.4692675173 &$\pm$& \num{2.2929864170312996e-3} & 0.37871593 &$\pm$& \num{1.108126258972789e-3} & 1.3722789 &$\pm$& \num{9.16861283418765e-3} \\ 
    {dG-VAE (ours)} &   0.018977943807840347 &$\pm$& \num{1.1629037296320311e-4} &  0.0112340897 &$\pm$& \num{5.7067902420742375e-05} & \B 0.0479579382 &$\pm$& \num{3.50237525617738e-4} & \B 0.30214578 &$\pm$& \num{1.0205811622208104e-3} & 1.0631448 &$\pm$& \num{6.8240103906364435e-3} \\ 
    {dG-VAE + restricted lab.}  & \B 0.015608689747750759  &$\pm$& \num{7.247633727085565e-5} & \B 0.0099587934 &$\pm$& \num{3.6474369756560165e-05} & 0.0630816892 &$\pm$& \num{3.4878623095947986e-4} &  0.308557845 &$\pm$& \num{1.348714427901733e-3} & \B 1.03199444 &$\pm$& \num{1.086903787052776e-2} \\ 
    \bottomrule
  \end{tabular}
\end{table*}

\cref{fig:spring_performance_samples} shows trajectory examples generated by each model. While the green point mass is headed upwards, all other predictions cluster in the lower half of the visualization. We can clearly see an improvement when comparing our method to other baselines, the predictions are much cleaner and closer to the ground truth than dNRI and LSTM. This observation is also supported by our numerical results in \cref{tab:spring_performance_comparison}, where our method outperforms all other approaches by almost an order of magnitude.  We note that the addition of restricted labeling yields an additional  5\% improvement in RMSE and ADE and, more importantly, improves the graph accuracy to near perfect: dNRI has an accuracy of 0.877 and dG-VAE with restricted labeling has 0.984.

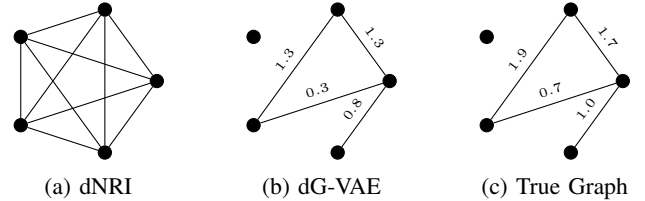
\begin{figure}[!tbp]
    \centering
    \begin{subfigure}[b]{0.3\linewidth}
        \centering
        \begin{tikzpicture}[scale=0.4]
              \def\numNodes{5}
              \def\radius{2.5cm}
            
              \foreach \s in {1,...,\numNodes} {
                \node[circle, draw, fill=black, inner sep=0pt, minimum size=5pt] (node\s) at ({360/\numNodes * (\s - 1)}:\radius) {};
              }
              \foreach \s in {1,...,\numNodes} {
                \pgfmathtruncatemacro{\nextNode}{mod(\s, \numNodes) + 1}
                \draw (node\s) -- (node\nextNode);
              }
              \draw (node1) -- (node3);
              \draw (node1) -- (node4);
              \draw (node2) -- (node5);
              \draw (node3) -- (node5);
              \draw (node4) -- (node2);
            \end{tikzpicture}
        \caption{dNRI}
    \end{subfigure}
    \hfill
    \begin{subfigure}[b]{0.3\linewidth}
        \centering
        \begin{tikzpicture}[scale=0.4]
              \def\numNodes{5}
              \def\radius{2.5cm}
            
              \foreach \s in {1,...,\numNodes} {
                \node[circle, draw, fill=black, inner sep=0pt, minimum size=5pt] (node\s) at ({360/\numNodes * (\s - 1)}:\radius) {};
              }
            
              \draw (node2) -- (node4) node [midway,above,sloped,font=\tiny] {$1.3$};
              \draw (node4) -- (node1) node [midway,above,sloped,font=\tiny] {$0.3$};
              \draw (node2) -- (node1) node [midway,above,sloped,font=\tiny] {$1.3$};
              \draw (node1) -- (node5) node [midway,above,sloped,font=\tiny] {$0.8$};
        \end{tikzpicture}
        \caption{\ours}
    \end{subfigure}
    \hfill
    \begin{subfigure}[b]{0.3\linewidth}
        \centering
        \begin{tikzpicture}[scale=0.4]
              \def\numNodes{5}
              \def\radius{2.5cm}
            
              \foreach \s in {1,...,\numNodes} {
                \node[circle, draw, fill=black, inner sep=0pt, minimum size=5pt] (node\s) at ({360/\numNodes * (\s - 1)}:\radius) {};
              }
            
              \draw (node2) -- (node4) node [midway,above,sloped,font=\tiny] {$1.9$};
              \draw (node4) -- (node1) node [midway,above,sloped,font=\tiny] {$0.7$};
              \draw (node2) -- (node1) node [midway,above,sloped,font=\tiny] {$1.7$};
              \draw (node1) -- (node5) node [midway,above,sloped,font=\tiny] {$1.0$};
        \end{tikzpicture}
        \caption{True Graph}
    \end{subfigure}
    \caption{Comparison of inferred graphs for dNRI and \ours. The edge labels refer to spring constants $k$, which \ours ~is learning through restricted labeling.}
    \label{fig:spring_inferred_graphs}
\vspace{-0.4cm}
\end{figure}

\cref{fig:spring_inferred_graphs} provides the inferred graphs for the same sample scenario.
While dNRI infers a fully connected graph, \ours ~shows a more adequate inference, such as the green node's predicted trajectory to be independent of any other point mass. Restricted labeling leads to an additional improvement by inferring spring constants. While our method's predictions are promising, inferring stronger spring constants where needed, they could be improved further.

\subsection{Motion Capture}\label{ssec:motion}

We also evaluate the efficacy of our proposed method using motion capture recordings sourced from the CMU Motion Capture database.\footnote{\url{http://mocap.cs.cmu.edu}} These recordings were obtained through a motion capture system featuring 12 infrared cameras at a frequency of 120 Hz. The dataset includes recordings that track movements across 31 distinct joints of the subject's body. Specifically, we analyze sequence 35, which involves walking, and use sequence 39, which tracks another test subject walking on uneven grounds, as OOD samples. It is difficult to define OOD for any application, but, for the purpose of this paper, we consider these sequences to provide data for evaluating the effectiveness of our method in practical motion analysis.

\begin{figure}[t]
    \begin{subfigure}[b]{0.3\linewidth}
        \centering
        \hspace*{-0.0cm}\includegraphics[width=\linewidth]{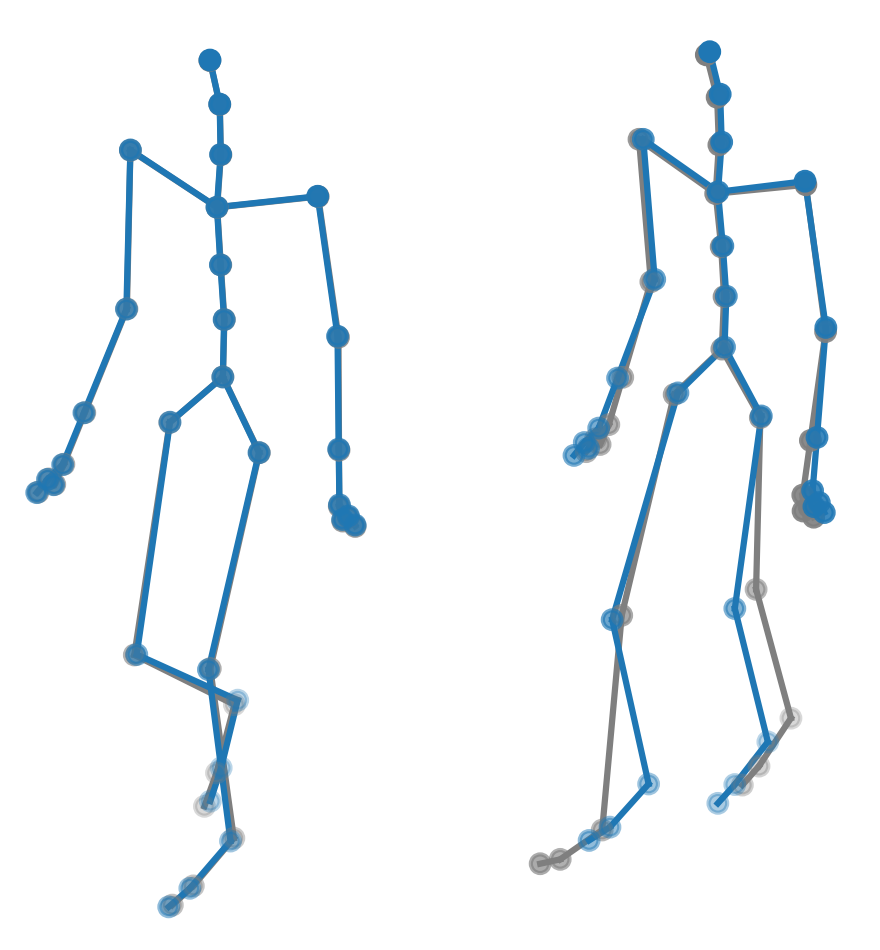}
        \caption{dNRI}
    \end{subfigure}
    \hfill
    \begin{subfigure}[b]{0.3\linewidth}
        \centering
        \hspace*{-0.0cm}\includegraphics[width=\linewidth]{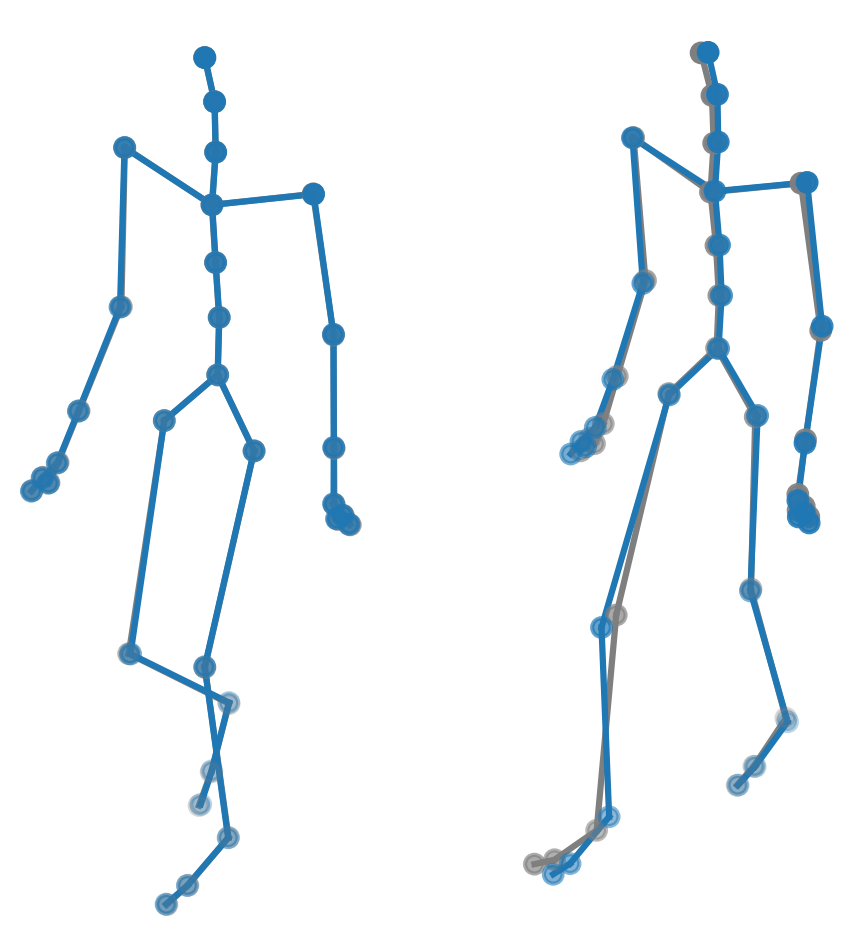}
        \caption{dG-VAE}
    \end{subfigure}
    \hfill
    \begin{subfigure}[b]{0.3\linewidth}
        \centering
        \hspace*{-0.0cm}\includegraphics[width=\linewidth]{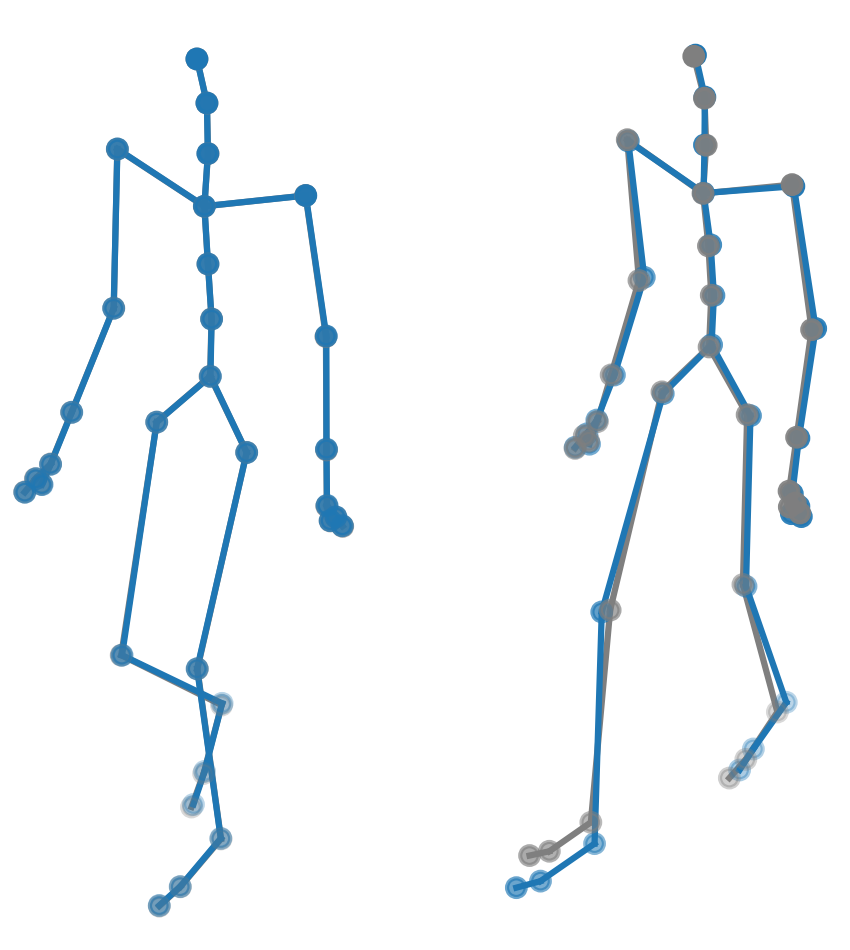}
        \caption{LSTM}
    \end{subfigure}
    \caption{(Motion Capture \#35) Samples from prediction models at time steps 25 and 50. The grey skeletons represent the ground truth, and the blue ones are the predictions.}
    \label{fig:motion_performance_samples}
\end{figure}

In a comparative overview of our method in \cref{tab:motion_performance_comparison}, we notice that it significantly outperforms dNRI on all three metrics, with a 45\% improvement. These findings are also reflected qualitatively in \cref{fig:motion_performance_samples}, where \ours ~matched the longer prediction (i.e., at $t=50$) much closer than dNRI and LSTM, especially around the legs and feet.

\begin{table*}[h]
  \caption{Performance comparison for Motion Capture (\#35) dataset}
  \label{tab:motion_performance_comparison}
  \centering
  \begin{tabular}{@{}p{30mm}|S[table-column-width=8ex, round-precision=3]@{\,} c @{\,}S[round-precision=2, table-column-width=9ex]S[table-column-width=8ex, round-precision=3]@{\,} c @{\,}S[round-precision=2, table-column-width=9ex]S[table-column-width=8ex, round-precision=3]@{\,} c @{\,}S[round-precision=2, table-column-width=10ex]|S[table-column-width=8ex, round-precision=3]@{\,} c @{\,}S[round-precision=2, table-column-width=10ex]S[table-column-width=8ex, round-precision=3]@{\,} c @{\,}S[round-precision=2, table-column-width=10ex]@{}}
    \toprule
    & \multicolumn{9}{c}{\#35} & \multicolumn{6}{c}{\#39}\\
    {Model} & \multicolumn{3}{c}{RMSE [m]} & \multicolumn{3}{c}{min ADE [m]}  & \multicolumn{3}{c|}{min FDE [m]} & \multicolumn{3}{c}{min ADE [m]} & \multicolumn{3}{c}{min FDE [m]} \\ 
    \midrule
    {Linear} & 2.7258792 &$\pm$ & \num{4.7361744383152424e-2} & 3.9890466 &$\pm$ & \num{6.164856139012265e-2} & 4.4931254 &$\pm$ & \num{1.2513358841685446e-1} & 5.4658999443 &$\pm$ & \num{2.9073837e-1} & 5.7503261566 & $\pm$ & \num{6.554395037391972e-1} \\
    {LSTM} & 0.20821631 &$\pm$ & \num{7.036836952505876e-3} & 0.16715486 &$\pm$ & \num{4.768830546501608e-3} & 0.82940567 &$\pm$ & \num{2.849375595547026e-2} & 0.29073837 & $\pm$ & \num{1.8589759685414688e-2} & 1.39768 & $\pm$ & \num{1.0673901734634755e-1}\\
    {IMMA} & 0.41836205 &$\pm$ & \num{9.993972290566386e-3} & 0.58260214 &$\pm$ & \num{1.3586789599648177e-1} & 1.081191 &$\pm$ & \num{2.4518406279579744e-1} & 1.2739396 &$\pm$ & \num{2.0018030217196045e-2} & 2.327761&$\pm$ & \num{3.238529647818474e-2}\\ 
    {dNRI} & 0.16102853 &$\pm$ & \num{2.5970360840623565e-3} & 0.14493029 &$\pm$ & \num{1.1882505731372554e-3} & 0.6992854 &$\pm$ & \num{1.2880715110425043e-2} & 0.2276797444 & $\pm$ & \num{5.890292930975932e-3} & 1.1241989136 & $\pm$ & \num{6.875674564569896e-2}\\
    {dG-VAE (ours)} & \B 0.06814845 &$\pm$ & \num{1.3338381890919547e-3} & \B 0.07476496 &$\pm$ & \num{1.3580907826643597e-3} & \B 0.29379427 &$\pm$ & \num{5.67878375897709e-3} & 0.1684396118 & $\pm$ & \num{8.849477116412796e-3} & 0.7900165319 & $\pm$ & \num{4.448753371861913e-2}\\
    {dG-VAE + pair matching} & 0.088167764 &$\pm$ & \num{3.7989371361645766e-3} & 0.082526386 &$\pm$ & \num{2.0364608711190726e-3} & 0.34435076 &$\pm$ & \num{1.115468364688169e-2} & \B 0.1674835736 & $\pm$ & \num{6.892894095513857e-3} & \B 0.7864799738 & $\pm$ & \num{4.913255294129185e-2}\\
    \bottomrule
  \end{tabular}
\end{table*}

\begin{table*}[h]
  \caption{Performance comparison for inD dataset}
  \label{tab:ind_performance_comparison}
  \centering
  \begin{tabular}{@{}p{30mm}|S[table-column-width=8ex, round-precision=3]@{\,} c @{\,}S[round-precision=2, table-column-width=9ex]S[table-column-width=8ex, round-precision=3]@{\,} c @{\,}S[round-precision=2, table-column-width=9ex]S[table-column-width=8ex, round-precision=3]@{\,} c @{\,}S[round-precision=2, table-column-width=10ex]|S[table-column-width=10ex, round-precision=3]@{\,} c @{\,}S[round-precision=3, table-column-width=8ex]S[table-column-width=10ex, round-precision=3]@{\,} c @{\,}S[round-precision=3, table-column-width=8ex]@{}}
    \toprule
    & \multicolumn{9}{c}{inD} & \multicolumn{6}{c}{rounD}\\
    {Model} & \multicolumn{3}{c}{RMSE [m]} & \multicolumn{3}{c}{min ADE [m]}  & \multicolumn{3}{c|}{min FDE [m]} & \multicolumn{3}{c}{min ADE [m]} & \multicolumn{3}{c}{min FDE [m]} \\ 
    \midrule
    dNRI  &  1.0798777 &$\pm$ & \num{3.140053692916578e-2} & 0.5614866 &$\pm$ & \num{3.141483121318685e-2} & 4.290376 & $\pm$ & \num{1.6744488236728386e-1} & 4.830788 & $\pm$ & 0.08596033114831662 & 37.07748 & $\pm$ & 0.9655629397674838 \\
    dG-VAE (ours)  &  \B 1.056488  &$\pm$ & \num{2.8273907201625443e-2} & \B 0.48634544 &$\pm$ & \num{1.4381765705679886e-2} & \B 4.284938 &$\pm$ & \num{1.03760481732467e-1} & 4.95847623 & $\pm$ & 0.07588964458152 & 38.98547 & $\pm$ & 1.14576359634857 \\
    dG-VAE + pair matching &  1.1044311571121216 &$\pm$ & \num{5.5473567305058515e-2} & 0.5576806068 &$\pm$ & \num{1.5671481642484976e-2} & 4.4405803680 &$\pm$ & \num{3.3611905983237605e-1} & \B 4.384334 & $\pm$ & 0.11081128897601174 & \B 34.84829 & $\pm$ & 1.2791184085258698\\
    \bottomrule
  \end{tabular}
\end{table*}

\begin{figure}[t]
    \begin{center}
    \hspace*{-0.4cm}\includegraphics[width=0.8\linewidth]{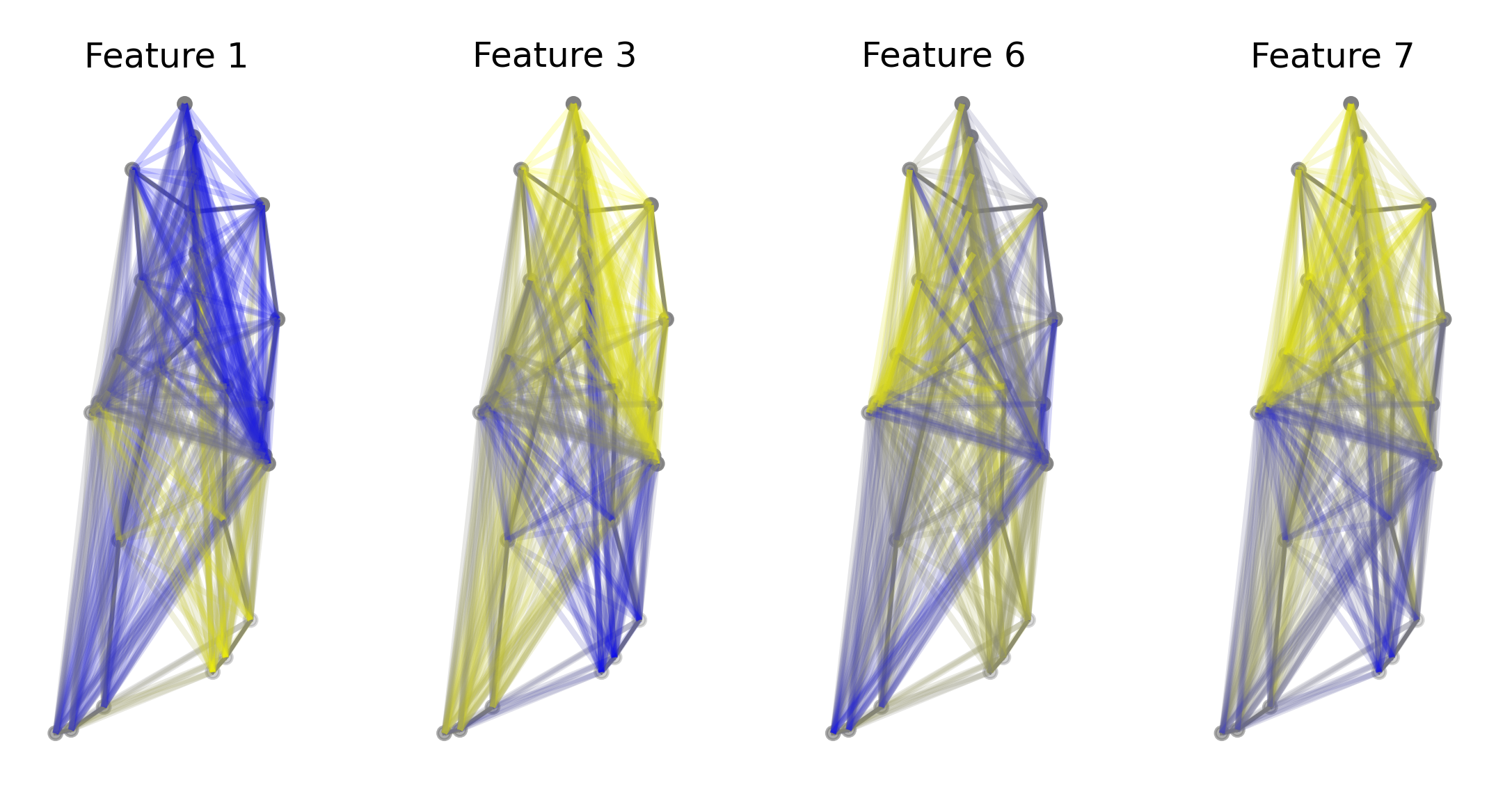}
    \hspace*{0.4cm}\raisebox{0.35cm}{\includegraphics[height=0.28\linewidth]{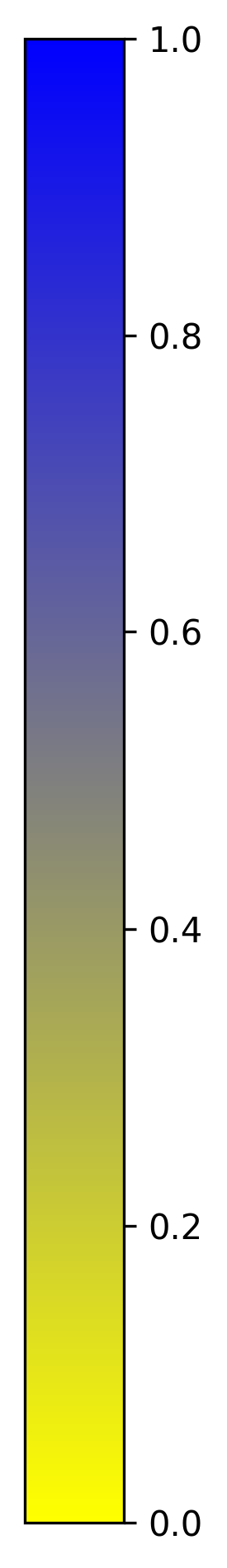}}
    \end{center}
    \caption{A subset of stationary features learned by dG-VAE.}
    \label{fig:motion_stationary_features}
\vspace{-0.2cm}
\end{figure}

While disentanglement appears to not significantly affect performance quantitatively as seen in the previous two test cases, it introduces a new level of expressivity, as illustrated in \cref{fig:motion_stationary_features}. We first note that a very limited number of features are associated with nodes along the centerline, such as the core, shoulders, and head. This minimal correlation indicates that other factors play a more significant role in predicting future motion. 
Overall, the learned features seem to correlate with relatively distinct relations:
\begin{enumerate}
    \item Feature 1 showcases the main relation of the upper body to the front leg, which suggests this feature might encode weight shifting when a step is taken.
    \item Feature 3 encodes the relation from the left foot and knee to the right hand and shoulder, and feature 6 encodes the opposite, i.e., the correlation between the right foot and the left hand and shoulder. These features highlight a common walking pattern where our arms swing synchronously with the opposite leg. 
    \item Feature 7 strongly relates all extremities, i.e., hands and feet, and weakly links more stationary relations. This encoding, therefore, suggests a variance in distance, e.g., while the distance between shoulders remains relatively constant, the distance between feet increases and decreases like a pendulum as each step is taken.
\end{enumerate}

The features discussed here are only a subset to show how feature matrices can be interpreted like filters. We refer the reader to \cref{app:1} for a more complete overview of static and dynamic edge features.

\subsection{inD}\label{ssec:ind}

The {inD} dataset \citep{inD} is composed of vehicle tracks extracted from 33 drone recordings at four German intersections. For each track, metadata such as agent type (i.e., pedestrian, car, truck, bike), width, height, and the number of frames it is present for, is given in addition to its trajectory. 
The feature vector comprises $x,y$ position, heading, and lateral and longitudinal velocity and acceleration. The positional error from the drone recordings is expected to be less than $\qty[round-precision=1]{10}{\cm}$. 
The OOD generalization is measured by deploying a model trained on inD data and evaluating its performance on the rounD dataset. As the name suggests, the rounD dataset is a sister dataset of inD but for recordings in roundabouts. Frames were taken at a $\qty[round-precision=1]{0.2}{\second}$ interval, resulting in a $\qty[round-precision=1]{4}{\second}$ prediction horizon when evaluating $T=20$ steps.

\begin{figure}[t]
    \begin{subfigure}[b]{0.5\linewidth}
        \hspace*{-0.5cm}\includegraphics[height=\linewidth]{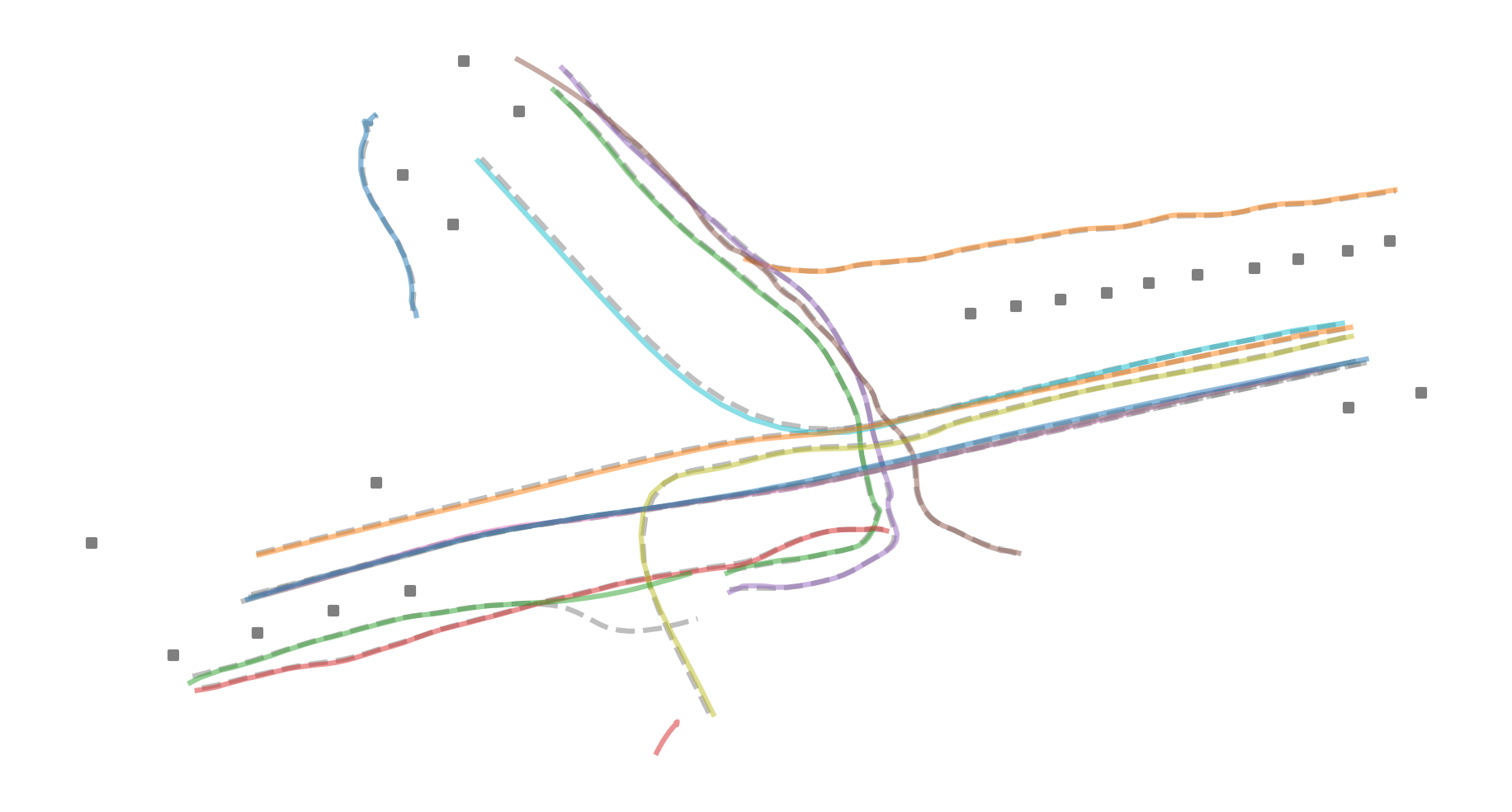}
        \caption{Predicted trajectories.}
        \label{fig:ind_performance_samples_a}
    \end{subfigure}
    \hfill
    \begin{subfigure}[b]{0.45\linewidth}
        \centering
        \includegraphics[height=0.5\linewidth]{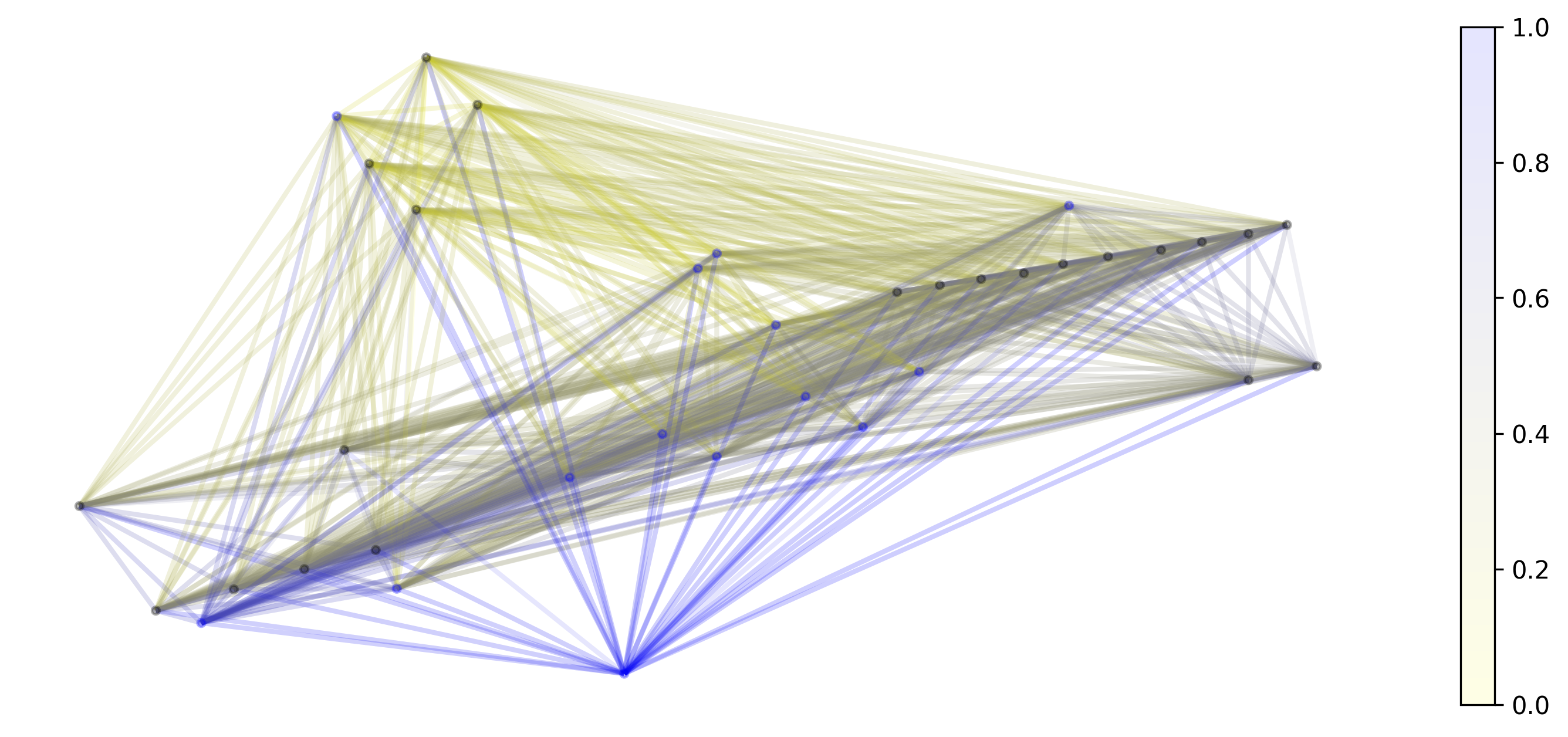}
        \caption{Edge embeddings.}
        \label{fig:ind_performance_samples_b}
    \end{subfigure}
    \caption{(inD) (a) Trajectories sampled from dG-VAE. Grey lines represent ground truth trajectories and squares are stationary cars. (b) Edge embeddings at $t=50$.}
    \label{fig:ind_performance_samples}
    \vspace{-0.2cm}
\end{figure}

The experimental results summarized in \cref{tab:ind_performance_comparison} highlight that while learning edge features produces significant improvements on smaller graphs, e.g., the Spring dataset with 5 nodes sees a performance increase of an order of magnitude and the Motion dataset featuring 31 nodes sees around 42\%, the return is diminished on larger ones: inD with 500 nodes on average sees around 5-10\% improvement. Similarly, disentangling methods have different advantages depending on the test case. While always improving interpretability and OOD generalizability, the improvements in distribution, beyond feature learning, are limited in specific test cases. 

\cref{fig:ind_performance_samples_a} shows trajectory examples generated using dG-VAE. The trajectories are well matched with a few exceptions, e.g., the light green and purple trajectories deviate slightly towards the end. \cref{fig:ind_performance_samples_b} is a visualization of the embedded edge features learned by dG-VAE at the same time step. What makes this intriguing is how the model discerns between edges that connect two moving vehicles and edges linking two stationary entities, attributing high-value features to the former and near-zero feature weights to the latter. Edges that link stationary and moving agents demonstrate feature weights that lie somewhere in between. From this, we can infer that the network can successfully recognize connections significant for predicting the future positions of agents. 

\textbf{Limitations.} Our implementation of dG-VAE has some limitations, in terms of data dependency, model generalization, and interpretability. The model generally requires large amounts of high-fidelity data for learning trends in latent relations. Generalizability in our model requires that the underlying dynamics remain unchanged, e.g., rules of the roads still apply. 
Although our model was designed for interpretability, additional postprocessing may be required to make the inferences more human-understandable.

\section{Conclusion}
Our work addresses the critical challenges of interpretability and out-of-distribution generalizability in the context of interaction modeling and behavior prediction for dynamic agents. We designed a variational auto-encoder framework that integrates graph-based representations and recurrent neural networks, enabling efficient capture of spatio-temporal relations and higher prediction accuracy. Our model uses a latent space that infers dynamic interaction graphs, enriched with interpretable edge features characterizing the interactions and identifying latent factors. Furthermore, we employed two techniques to disentangle the latent space of edge features, thereby enhancing model interpretability and performance in out-of-distribution scenarios.

We demonstrate the effectiveness of learning edge features on multiple datasets, surpassing existing methods in predicting future interactions and motions. While feature space disentanglement further improved performance on smaller graphs, e.g., NBA and Spring datasets saw an additional 20\% improvement, these benefits were limited on larger ones. Disentangling still yielded some gains in OOD generalization and, most importantly, created directly interpretable or latent static features that were correlated to semantically meaningful relations. In future work, we will explore possible interpretations of these disentangled embeddings.

\vfill
\textbf{Acknowledgments.} This work is supported by Honda Research Institute USA.

\appendix

\section{}\label{app:1}

\vspace{-0.2cm}

\begin{figure}[H]
    \captionsetup{font=small}
    \begin{subfigure}[b]{0.15\linewidth}
        \centering
        \hspace*{-0.0cm}\includegraphics[height=3.5cm]{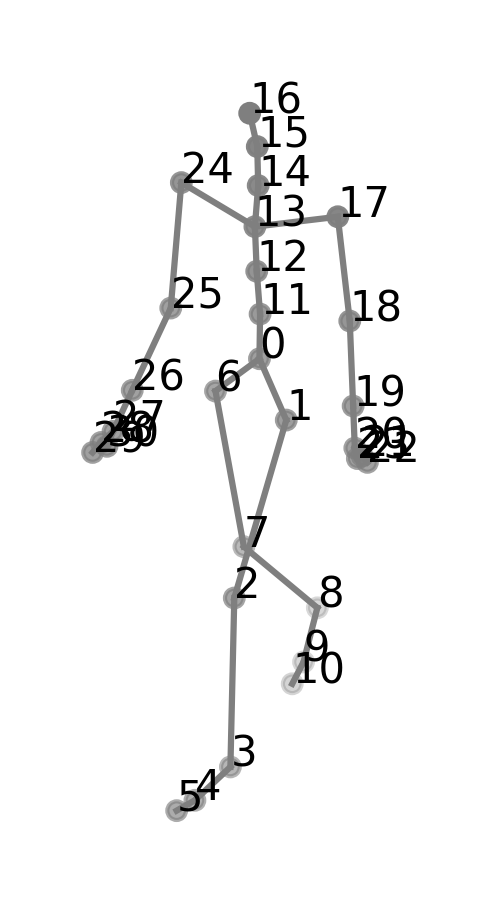}
    \end{subfigure}
    \hfill
    \begin{subfigure}[b]{0.74\linewidth}
        \centering
        \hspace*{-0.5cm}\includegraphics[height=3.5cm]{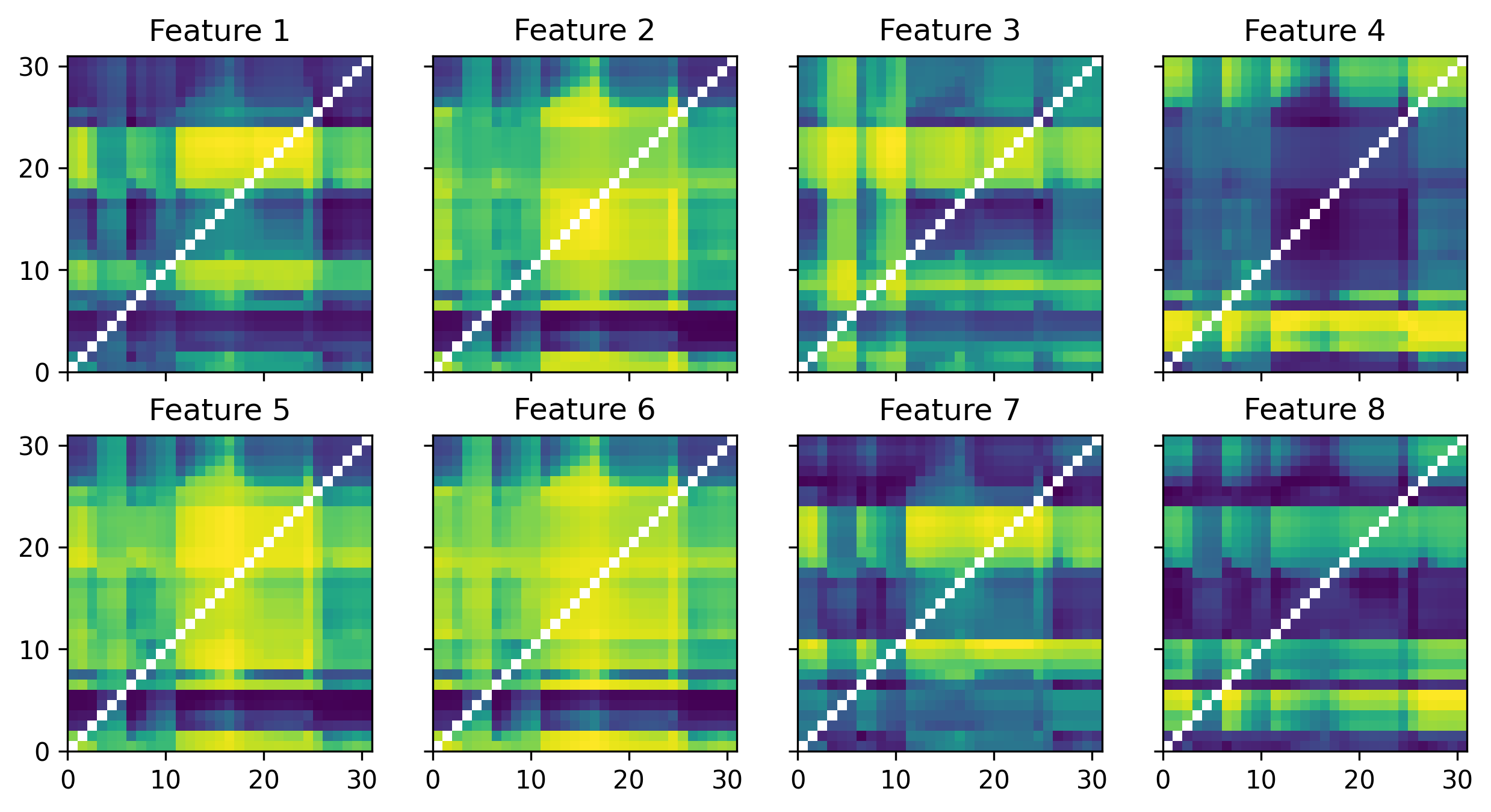}
    \end{subfigure}
    \caption{Stationary features learned through pair matching.}
    ~\\
    \begin{subfigure}[b]{0.15\linewidth}
        \centering
        \hspace*{-0.0cm}\includegraphics[height=3.5cm]{ieeeconf/figs/gt_0.png}
    \end{subfigure}
    \hfill
    \begin{subfigure}[b]{0.74\linewidth}
        \centering
        \hspace*{-0.5cm}\includegraphics[height=3.5cm]{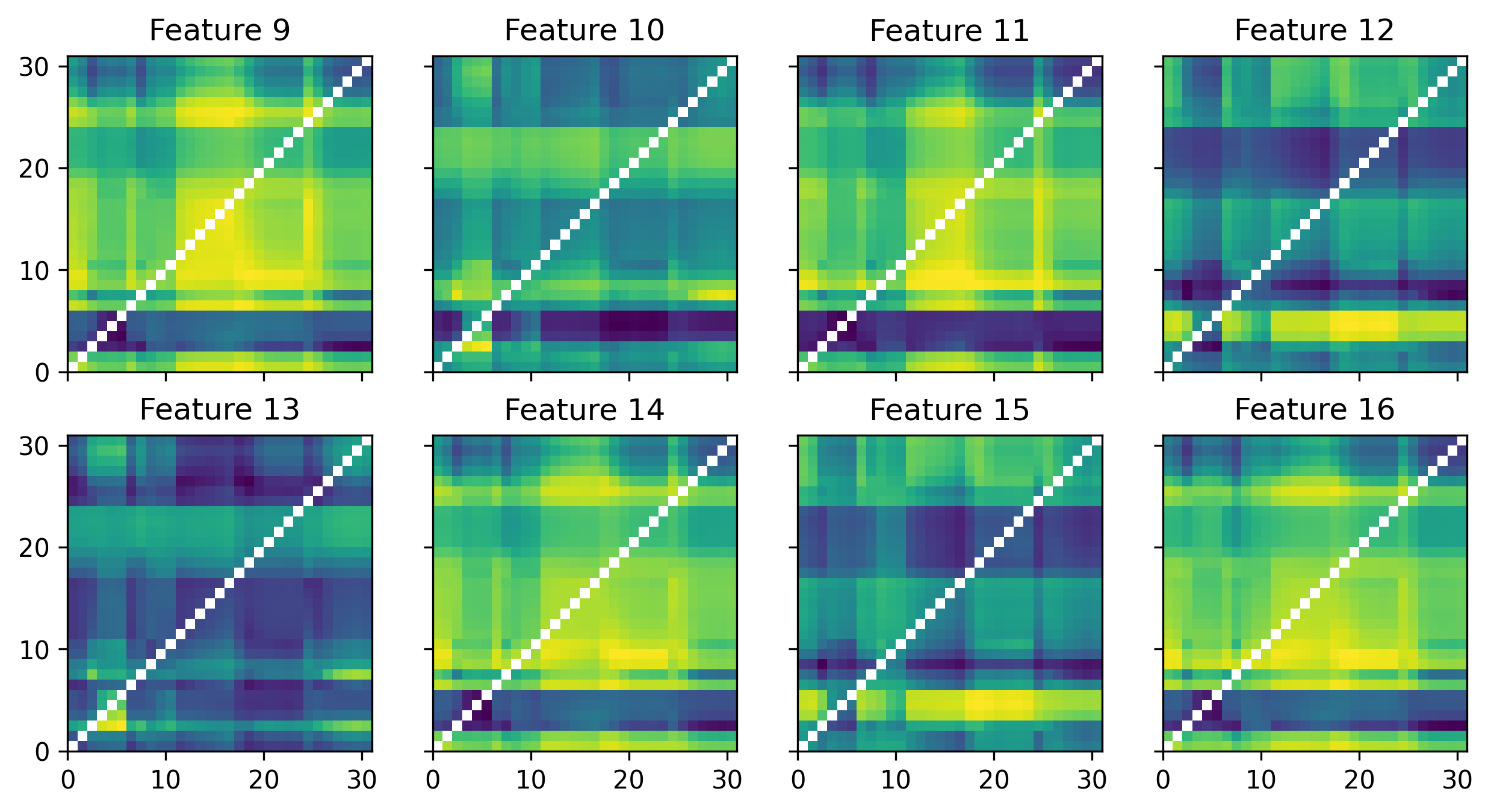}
    \end{subfigure}
    \caption{Dynamic features learned with \ours ~at $t=0$.}
    ~\\
    \begin{subfigure}[b]{0.15\linewidth}
        \centering
        \hspace*{-0.0cm}\includegraphics[height=3.5cm]{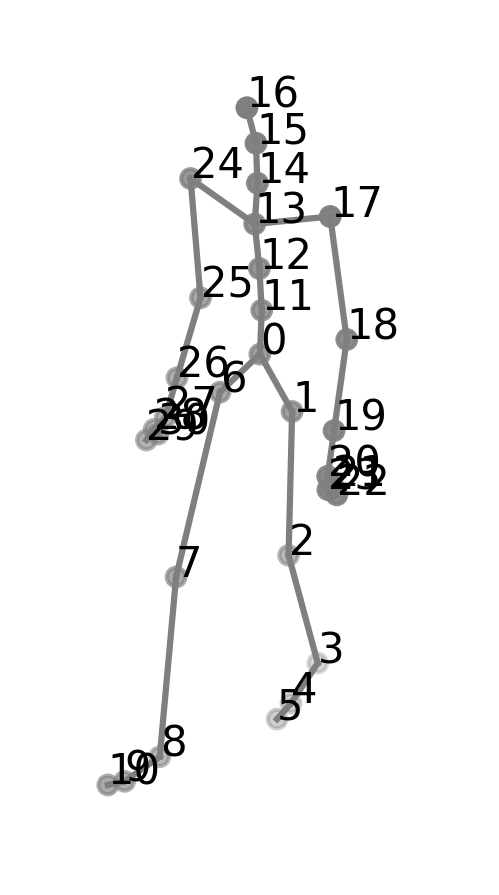}
    \end{subfigure}
    \hfill
    \begin{subfigure}[b]{0.74\linewidth}
        \centering
        \hspace*{-0.5cm}\includegraphics[height=3.5cm]{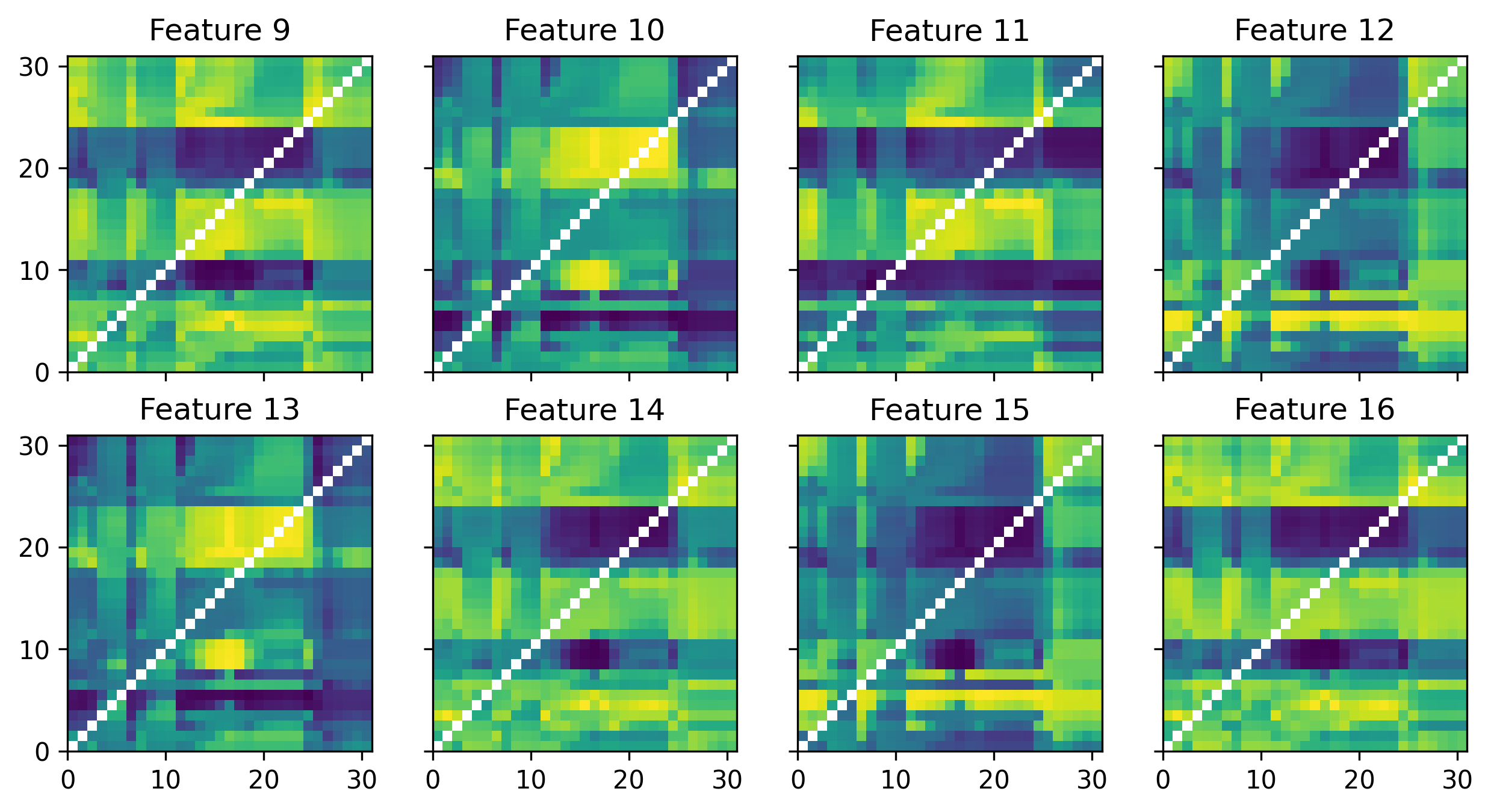}
    \end{subfigure}
    \caption{Dynamic features learned with \ours ~at $t=50$.}
    \vspace{-0.2cm}
    \label{fig:motion_feature_matrix}
\end{figure}

\renewcommand*{\bibfont}{\footnotesize}
\printbibliography

\end{document}